\documentclass{article}

 \usepackage[preprint]{neurips_2025}

\usepackage[utf8]{inputenc} 
\usepackage[T1]{fontenc}    
\usepackage{hyperref}       
\usepackage{url}            
\usepackage{booktabs}       
\usepackage{amsfonts}       
\usepackage{nicefrac}       
\usepackage{microtype}      
\usepackage{xcolor}         
\usepackage{graphicx}
\usepackage{natbib} 

\usepackage{amsmath}
\usepackage{cleveref}
\usepackage{chngcntr}

\usepackage{subcaption} 
\usepackage{enumitem}
 
\usepackage{algorithm} 
\usepackage{algorithmic}

\title{From Points to Spheres: A Geometric Reinterpretation of Variational Autoencoders}

\author{Songxuan Shi \\
Department of Applied Physics \\
Beijing University of Technology \\
Beijing 100124, China \\
\texttt{shisongxuan@emails.bjut.edu.cn} \\
}

\begin{document}

\maketitle

\begin{abstract}
Variational Autoencoder is typically understood from the perspective of probabilistic inference. In this work, we propose a new geometric reinterpretation which complements the probabilistic view and enhances its intuitiveness. We demonstrate that the proper construction of semantic manifolds arises primarily from the constraining effect of the KL divergence on the encoder. We view the latent representations as a Gaussian ball rather than deterministic points. Under the constraint of KL divergence, Gaussian ball regularizes the latent space, promoting a more uniform distribution of encodings. Furthermore, we show that reparameterization establishes a critical contractual mechanism between the encoder and decoder, enabling the decoder to learn how to reconstruct from these stochastic regions. We further connect this viewpoint with VQ-VAE, offering a unified perspective: VQ-VAE can be seen as an autoencoder where encodings are constrained to a set of cluster centers, with its generative capability arising from the compactness rather than its stochasticity. This geometric framework provides a new lens for understanding how VAE shapes the latent geometry to enable effective generation.
\end{abstract}

\section{Introduction}

\begin{figure}[h!]
    \centering
    \includegraphics[width=1\textwidth]{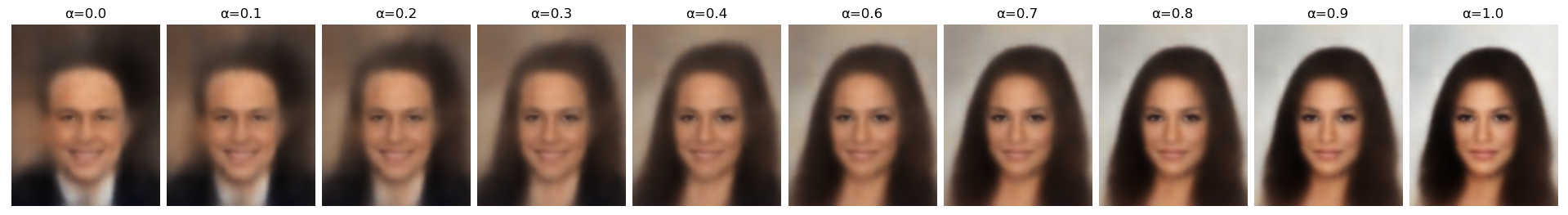} 
    \caption {\textbf{10-step linear interpolations in the latent space of Autoencoder}. We use a convolutional autoencoder with latent dimension of 786, trained on 10,000 images from the CelebA dataset for 10 epochs (batch size 32). The figure demonstrates that the model can achieve smooth and natural transitions through linear interpolation.} 
    \label{fig_AE_inter} 
\end{figure}

Autoencoder (AE)\cite{BP_origin_AE,AE02}, is a neural network composed of an encoder and a decoder. The encoder maps the input data \( \mathbf{x} \) to a latent representation \( \mathbf{z} \) through a network \( \mathbf{z} = f_{\theta}(\mathbf{x}) \). The decoder then reconstructs the input from the latent representation via \( \hat{\mathbf{x}} = g_{\phi}(\mathbf{z}) \). The goal of training an AE is to minimize the reconstruction loss, typically defined as \( \mathcal{L}(\theta, \phi) = \mathbb{E}[\|\mathbf{x} - g_{\phi}(f_{\theta}(\mathbf{x}))\|^2] \).

Variational Autoencoder (VAE)\cite{VAE} extends AE by introducing a probabilistic perspective on the latent space. The training objective of VAE is to maximize the evidence lower bound (ELBO), given by \( \mathcal{L}(\mathbf{x}) = \mathbb{E}_{q_{\phi}(\mathbf{z}|\mathbf{x})}[\log p_{\theta}(\mathbf{x}|\mathbf{z})] -  D_{KL} (q_{\phi}(\mathbf{z}|\mathbf{x}) \| p(\mathbf{z})) \), where \( p(\mathbf{z}) \) is a prior distribution (usually \( \mathbf{z} \sim \mathcal{N}(\mathbf{0}, \mathbf{I}) \)) and \( \text{KL} \) denotes the Kullback-Leibler divergence.

For a long time, we have been modeling VAE from a probabilistic perspective. 
The predecessor of VAE, AE, was long thought to lack generative capabilities. However, studies have achieved image generation using linear or circular interpolation with AE, as shown in \Cref{fig_AE_inter} (e.g. \( \mathbf{z}_{\alpha} = (1 - \alpha) \mathbf{z}_1 + \alpha \mathbf{z}_2 \)) \cite{AE_inter01}.
People then began to explain this using the language of manifold learning\cite{AE_manifold}: what we learn is a low-dimensional manifold of the data. \cite{HintonManifold} During interpolation, compared to random sampling, moving within the defined region of the manifold allows for smooth transitions. Nevertheless, in rare cases, the interpolation may unexpectedly move outside this region to undefined points, leading to breakdowns in AE interpolation.\cite{AE_inter_crash}
The use of interpolation for image generation is employed in multiple generative models. \cite{Gan_inter01,Gan_inter02}

However a question arises: if autoencoders can also exhibit generative capability, what exactly is their relationship with VAEs? What are the respective roles of the two key components in VAEs: KL divergence and reparameterization? Understanding VAEs through the lens of the variational lower bound fails to explain why standard AEs can also generate samples, which is puzzling and does not generalize to its discrete counterpart, VQ-VAE. These questions motivate our study. Our contributions can be summarized as follows:

\begin{enumerate}[
    topsep=0pt,  
    partopsep=0pt, 
    itemsep=2pt,  
    parsep=0pt,    
    leftmargin=*  
]
    \item Through empirical experiments, we identify \textbf{compactness of the latent space} as the fundamental factor enabling generative capability in autoencoding models.
    \item We clarify that \textbf{reparameterization} in VAE primarily serves as a regularization mechanism rather than being the direct source of generative ability.
    \item We propose two novel metrics for interpretable latent space analysis:
        \begin{enumerate}[
            topsep=0pt,    
            itemsep=1pt,   
            leftmargin=*]
            \item \textbf{Coefficient of Variation of Nearest-Neighbor Distances (CV-NND)}: measures the uniformity of local density in the latent space.
            \item \textbf{Dynamic Latent Coverage (DLC)}: evaluates how effectively the latent codes cover the space under dynamic sampling.
        \end{enumerate}
    \item We propose a \textbf{unified framework} in which both VQ-VAE and compactness-regularized AEs emerge as special cases. This model provides a coherent explanation for why VQ-VAE, despite lacking a continuous latent prior or reparameterization, still possesses strong generative capabilities.
\end{enumerate}

While some existing work explores VAE latent space using complex Riemannian geometry, we propose a more intuitive perspective based on empirical analysis rather than theoretical derivations. The observations and conclusions are new and distinct from those of traditional theoretical frameworks.

\section{Reparameterization as a Regularization Mechanism}

In this section, we aim to explore how the reparameterization in VAE can be interpreted through a non-probabilistic lens. We seek to understand why VAE are able to perform sampling-based generation, going beyond the mere interpolation seen in standard AE.

First, consider that in an AE, interpolated images, those lying between encoded points are not present in the training set\cite{AE_inter01}, yet they are meaningfully generated. This suggests that the neural network implicitly learns a semantic manifold. The internal dynamics of this manifold remain unknown, but its existence is evident from the smooth transitions observed in interpolation.
However, building a generative model solely as an AE is insufficient. The latent space tends to contain semantically void regions.\cite{AE_inter_crash} 

So, what does the VAE do differently to shape the latent space?

\subsection{Gaussian Balls Make the Latent Space More Permissive Under KL Divergence Constraint}

\begin{figure}[h!]
    \centering 
    \subfloat[]{\includegraphics[width=0.4\textwidth]{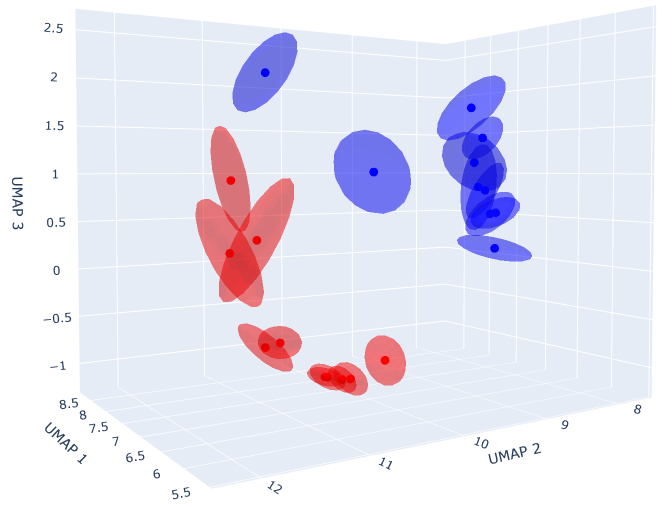}\label{fig:ellipsoid}}
    \hfill 
    \subfloat[]{\includegraphics[width=0.6\textwidth]{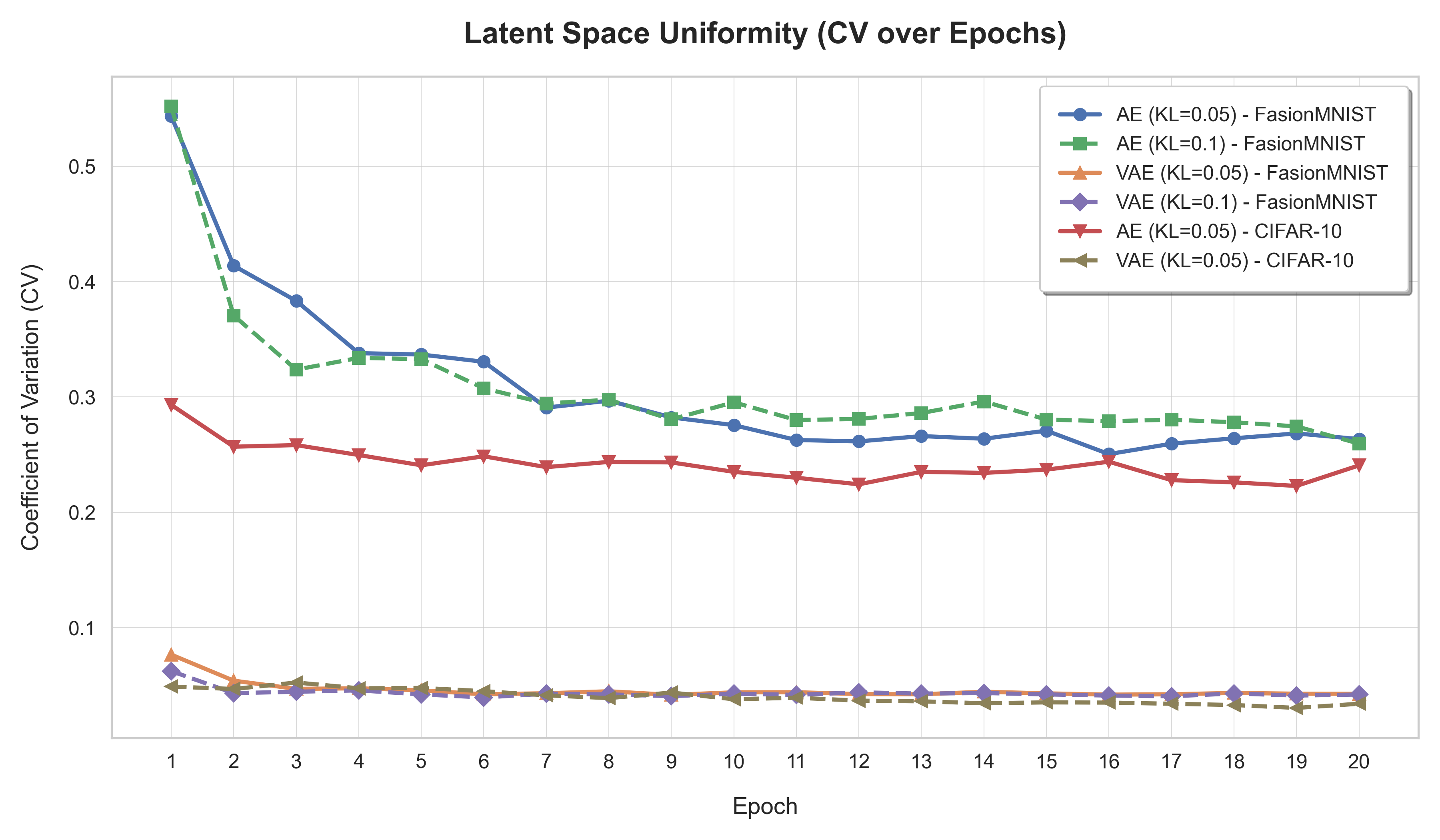}\label{fig:cv_plot}}
    \caption{
    \protect\subref{fig:ellipsoid}): Visualization of Gaussian ball in a VAE trained on MNIST. For 128 test samples, each encoded distribution (with mean \( \boldsymbol{\mu} \) and variance \( \boldsymbol{\sigma}^2 = \exp(\log\boldsymbol{\sigma}^2) \)) is interpreted as a high-dimensional ellipsoid. Blue and red ellipsoids correspond to 10 randomly selected samples from two distinct classes in the dataset. To visualize their geometric shapes in 3D, we use a \textit{sample-map-fit} procedure: (1) Draw samples from each latent Gaussian (2) Map these points to 3D using the UMAP method\cite{umap} (3) Fit an ellipsoid to the projected point cloud.
    \protect\subref{fig:cv_plot}):Coefficient of Variation (CV) of nearest-neighbor distances in the latent space, evaluated on CIFAR-10 and FashionMNIST under different \( \beta \)-KL constraints (0.05, 0.1) over 20 epochs. The evaluation was performed on a test set using a batch size of 128 across 16 batches, with each batch sampled 1,000 times.All models used Adam (lr=0.001), with an FFN for MNIST variants and a custom CNN for CIFAR-10.}
    \label{figCV_balls}
\end{figure}

In the reparameterization operation of VAE, we express the latent variable as:
\[
z = \mu + \sigma \odot \epsilon  \quad \epsilon \sim \mathcal{N}(0, I)
\]
where \( \mu \) and \( \sigma \) are the mean and standard deviation output by the encoder respectively.

We introduce the concept of a Gaussian ball, as shown in \Cref{fig:ellipsoid} (for visualization purposes only). Under reparameterization, each latent code is no longer a single point, but a probabilistic region, an \textbf{anisotropic ellipsoid} in latent space. This ellipsoid is centered at \( \hat{\mu} \), its shape determined by the variance vector \( \hat{\sigma} \).
We interpret this Gaussian ball as a zone in latent space where the model believes the true representation of the input likely resides. For a given encoding with probability density \( p \), the associated \( \hat{\sigma} \) defines the spatial extent of this belief.

\textbf{Because every training sample is encoded into such a Gaussian ball instead of a single point. Crucially, the KL divergence acts as an enforced constraint here. Without this mandatory regulation, these balls could arbitrarily expand or shrink, rendering the model no different from a standard AE. Instead, under the KL divergence's coercive effect, the balls collectively "push" against one another during training, they occupy stable volumes, preventing codes from collapsing into arbitrarily small regions.}

To evaluate whether the latent representations learned by VAE and AE are uniformly or unevenly distributed, we introduce a metric based on the Coefficient of Variation (CV) of nearest-neighbor distances in the latent space. For each encoded point \(\mathbf{z}_i\) in a test batch, we compute its squared Euclidean distance to the nearest other point in the same batch: \(d_i = \min_{j \neq i} \|\mathbf{z}_i - \mathbf{z}_j\|^2\). This yields a set of nearest-neighbor distances \(\{d_i\}_{i=1}^N\). We then compute the CV of this distribution as \(\text{CV} = \frac{s}{d}\), where \(s\) is the sample standard deviation and \(d\) is the mean of the \(d_i\). The CV is dimensionless and normalizes variability relative to the mean, enabling fair comparison across models with different latent space scales. A lower CV indicates more uniform distribution of points, while a higher CV suggests clustering or irregular spacing.

To ensure a fair comparison, we require the AE to also output both mean and variance, where the mean is passed through to the latent code while the variance is disregarded. We apply the same KL divergence loss as in VAE to enforce similar regularization, thus isolating the effect of stochastic sampling in the latent space.

As shown in \Cref{fig:cv_plot}, we can observe that, across different datasets and varying values of \( \beta \) as the KL divergence coefficient, the CV values for VAE are consistently lower than those for AE. This suggests that the Gaussian ball design in VAE encourages a more dispersed and geometrically regular latent representation, thereby reducing the fluctuation in nearest-neighbor Euclidean distances.

\subsection{Reformulated VAE Training by using Gaussian Ball assumption}

\begin{figure}[h!]
    \centering
    \subfloat[]{
        \includegraphics[width=1\textwidth]{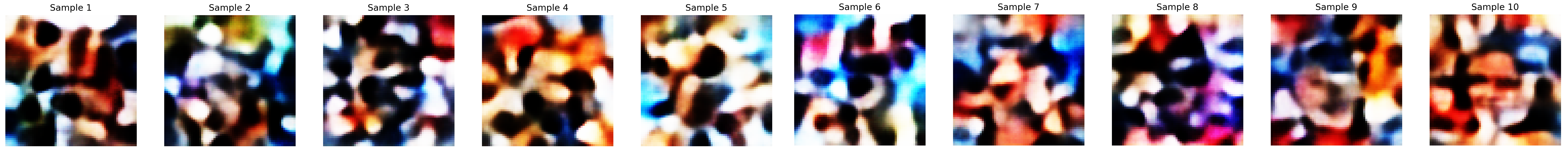}
        \label{fig:without_ball}
    }
    \\
    \subfloat[]{
        \includegraphics[width=1\textwidth]{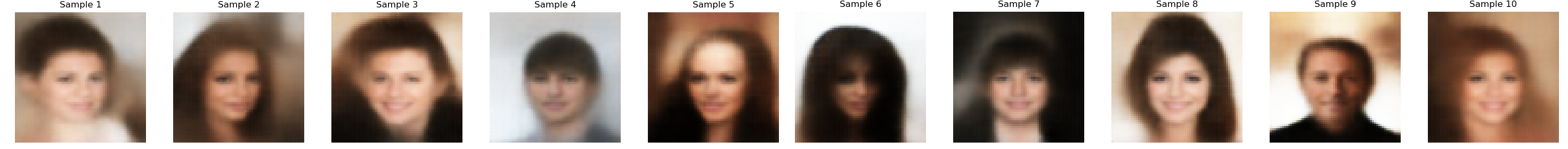}
        \label{fig:with_ball}
    }
    \caption{Comparison of random sampling results on CelebA. 
    Models are trained on \( 10^5 \) images using Adam ( \( \text{lr} = 0.001 \)) for 10 epochs. 
    The encoder is a convolutional network and the decoder uses transposed convolutions. 
    \protect\subref{fig:without_ball}: A baseline model without variance regularization. For a fair comparison, during generation, this model still employs reparameterization sampling, with the variance provided directly by the encoder's output.
     \protect\subref{fig:with_ball}: Our proposed asymmetric regularization method.}
    \label{fig:revised_vae}
\end{figure}

Based on the above analysis, we conclude that the reparameterization in VAE acts as a geometric regularizer that prevents the collapse of latent codes into deterministic points. To better understand its role, we decompose the regularization into two components: one controlling the location of the Gaussian ball (via \(\boldsymbol{\mu}\)), and another shaping its spatial extent (via the variance). This allows us to reformulate the training objective without relying on the full probabilistic framework of VAE.

We explore two variants of spread regularization that differ in their geometric inductive bias:

\textbf{Asymmetric Expansion}  
We penalize only under-dispersed Gaussian balls, allowing free expansion when radius is sufficiently large:
\[
\mathcal{L}_{\text{spread}}^{(1)} = \sum_i \mathrm{ReLU}(-\log\mathrm{Var}_i)
\]
Here, \(\log\mathrm{Var}_i\) denotes the log-variance output by the encoder for the \(i\)-th latent dimension identical to the standard VAE setup(not refer to sample-based estimate of variance). This term enforces a minimal scale: if \(\log\mathrm{Var}_i < 0\) (\(\mathrm{Var}_i < 1\)), a penalty is applied; otherwise, no constraint is imposed. This design reflects the principle: do not collapse, but grow freely, prioritizing diversity over uniformity.

\textbf{Symmetric Stabilization}  
Alternatively, we use:
\[
\mathcal{L}_{\text{spread}}^{(2)} = \sum_i \log\left(1 + (\log\mathrm{Var}_i)^2\right)
\]
which penalizes deviations of \(\log\mathrm{Var}_i\) from zero in both directions, encouraging the encoded variances to remain near unity. It's the same constraint as KL divergence, but with a different implementation.

In both cases, the full loss includes a centering term and reconstruction:
\[
\mathcal{L} = \underbrace{\ell_{\text{recon}}(x, \hat{x})}_{\text{reconstruction}} + \underbrace{\|\boldsymbol{\mu}\|^2}_{\text{centering}} + \underbrace{\mathcal{L}_{\text{spread}}}_{\text{spread regularization}}
\]

For comparison, we also write the standard VAE's KL divergence (with \(\mathcal{N}(0,I)\) prior):
\[
\mathcal{L}_{\text{KL}} = \frac{1}{2} \sum_i \left( \mu_i^2 + \sigma_i^2 - \log \sigma_i^2 - 1 \right), \quad \sigma_i^2 = \exp(\log\mathrm{Var}_i)
\]
This term simultaneously pulls \(\mu_i \to 0\) and \(\sigma_i^2 \to 1\).

Both versions can achieve the same results as VAE. Here, we only show the results for the Asymmetric Expansion version. As shown in \Cref{fig:with_ball}, models trained with this asymmetric regularization successfully generate meaningful samples via randomly sampling, demonstrating that the latent space remains sufficiently dispersed for stochastic decoding. In contrast, when the spread regularization is removed and only the centering term is kept, the latent codes collapse into a dense, fragmented structure (see \Cref{fig:without_ball}), failing to support stable sampling.

These results suggest that, we don't necessarily need pure KL divergence itself, which is mandated by the variational inference framework to achieve the same effect. The core function of the KL divergence is not to enforce a perfectly uniform or spherical latent geometry, but to \textbf{prevent the degeneration of Gaussian balls into point-like encodings}. Specifically, the KL divergence term, \( D_{KL}(q_\phi(z|x) || p(z)) \), where \( p(z) \) is a standard Gaussian with a variance of one, acts as a penalty that \textbf{pushes the variance of our learned Gaussian balls towards one}. This simple, yet crucial, mechanism ensures that the latent encodings occupy a meaningful volume rather than collapsing into singular points.

Once collapse is avoided, diverse and semantically meaningful generation becomes possible, even without strict scale control. This shifts the interpretation of VAE regularization from a probabilistic necessity to a geometric safeguard against representational collapse.

\subsection{Reparameterization as a Contractual Mechanism for Expanding the Set of Defined Points}

\begin{figure}[h!]
    \centering
    \subfloat[Additive perturbations on the first five dimensions of one encoded sample]{
        \includegraphics[width=0.45\textwidth]{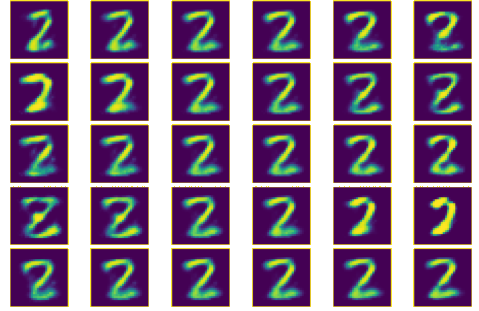}
        \label{fig:perturbation}
    }
    \hfill
    \subfloat[Reparameterized samples from five training inputs]{
        \includegraphics[width=0.43\textwidth]{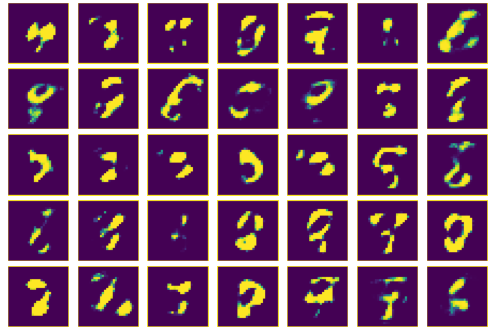}
        \label{fig:reparam_sampling}
    }
    
    \par\addvspace{10pt}
    
    \subfloat[A random sampling experiment on MNIST using an MLP, drawing from a Gaussian distribution.]{
        \includegraphics[width=0.6\textwidth]{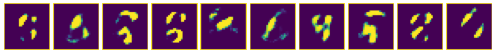}
        \label{fig:ramdom_sample_crash}
    }
    
    \caption{Decoder ablation experiment and random sampling on MNIST with an MLP architecture. Adam optimizer (lr = 0.001) is used for the ablation study.}
    \label{fig:combined_figures}
\end{figure}

In the previous section, we discussed how reparameterization in VAE introduces Gaussian balls that regularize the distribution of encoded points, promoting greater uniformity and preventing collapse. In contrast, AE typically maps inputs to deterministic points.

However, reparameterization does more than just regularize the geometry of the latent space. It effectively increases the points of semantic definition in the latent space. This means that during training, the model is exposed to a broader set of latent configurations, even from the same input.

This raises a critical question: Is the generative capability of VAE solely due to the regularization effect of these Gaussian balls, or does it also stem from the increased density of semantic definition, the fact that more regions of the space are actively used and optimized?

We design the following experiment to investigate the role of reparameterization in semantic coverage:

\begin{enumerate}
    \item Train a standard VAE with an MLP architecture on MNIST.
    \item Remove the trained decoder and replace it with a randomly initialized one.
    \item Freeze the encoder parameters and use the mean output \(\boldsymbol{\mu}\) as the latent code to train the new decoder, without any reparameterization or sampling.
\end{enumerate}

After training, we observe the following:
\begin{enumerate}
\item Reparameterization near the encoded vector (sampling from \(\mathcal{N}(\boldsymbol{\mu}, \boldsymbol{\sigma}^2)\)) fails to generate meaningful images, despite the original VAE having learned a well-structured latent space.\Cref{fig:perturbation}
    \item Sampling latent codes directly from a standard Gaussian \(\mathbf{z} \sim \mathcal{N}(0, I)\) also results in meaningless outputs, indicating that the new decoder does not inherit the prior compatibility of the original VAE. \Cref{fig:ramdom_sample_crash}
    \item Surprisingly, when we apply fixed additive perturbations along individual latent dimensions (e.g. \(z_i \leftarrow \mu_i + \delta\), for fixed \(\delta\)), smooth and semantically meaningful transitions emerge, despite the absence of stochastic sampling or KL regularization during training. \Cref{fig:reparam_sampling}
\end{enumerate}

While the underlying mechanism remains unclear, this result suggests that: \textbf{reparameterization and the KL divergence jointly regularize the geometry of the latent space. More importantly, the act of reparameterization itself effectively increases the density of semantic definition in the latent space. }

\section{Extending the Geometric View: Connection to VQ-VAE.}

\subsection{The Role of KL Divergence in Latent Space Filling.}

Previously, we discussed the synergistic effect of KL divergence and reparameterization in promoting latent space regularization. We then investigated the role of reparameterization itself, which increases the diversity of latent points and establishes a contract mechanism between encoder and decoder. In this section, we will examine the influence of KL divergence itself and its effect on filling the latent space.

We introduce \textit{Dynamic Latent Coverage}, a metric that tracks how well the encoder's representations cover the standard normal prior over the course of training. Inspired by prior work on representation sparsity~\cite{cheung2014discovering}, geometric structure in latent spaces~\cite{bengio2013representation}, and detection of underutilized regions in VAE~\cite{rubenstein2018learning}, our method provides a temporal view of latent space utilization by measuring the proportion of prior samples that fall within high-density neighborhoods of real data encodings.

The metric is computed at the end of each training epoch using a fixed evaluation set and an adaptive neighborhood radius. The procedure is given in \Cref{alg:dynamic_coverage}.

\begin{figure}[h!]
\centering
\begin{tabular}{ccc}
    \includegraphics[width=0.32\linewidth]{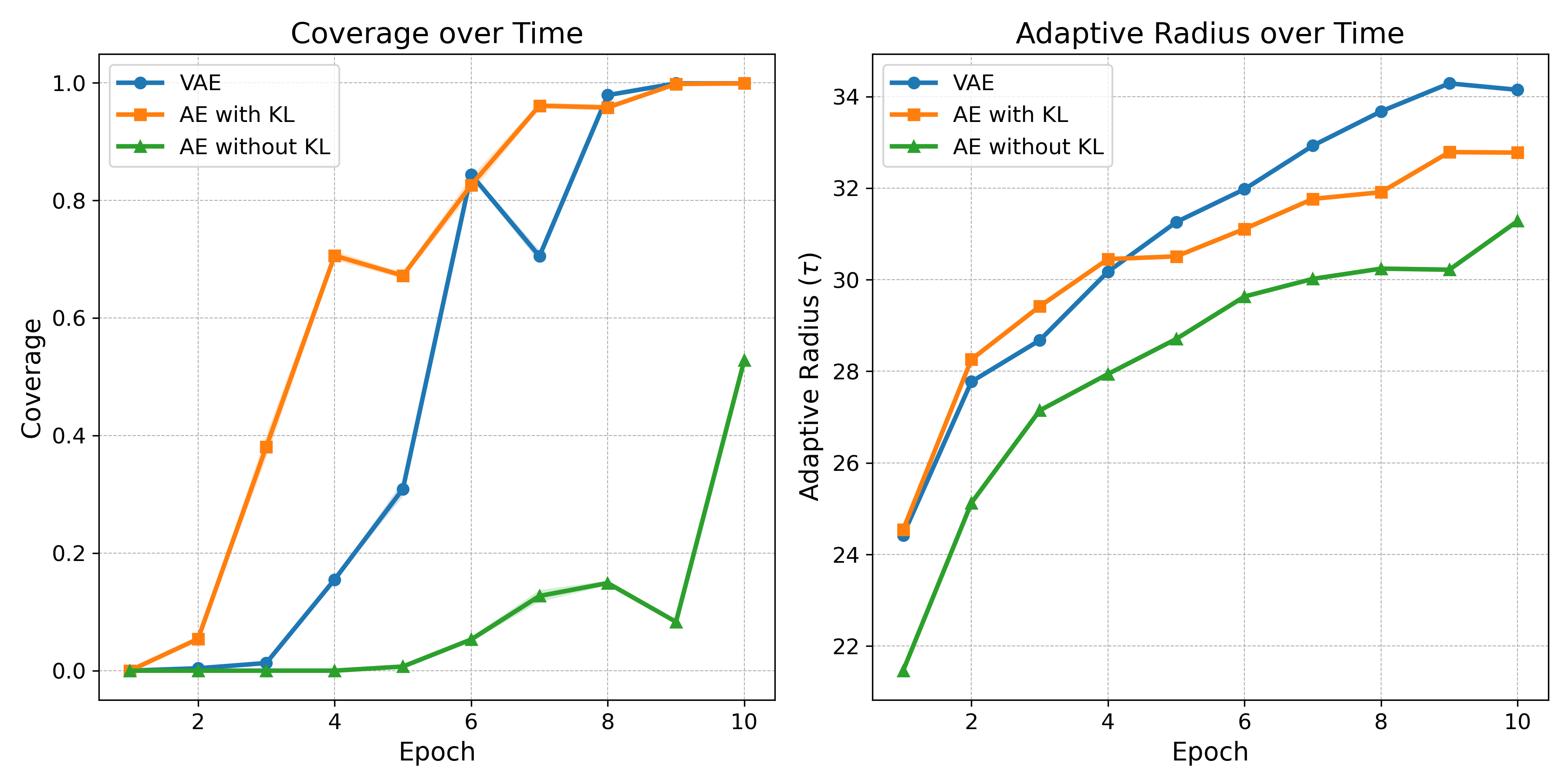} 
    \includegraphics[width=0.32\linewidth]{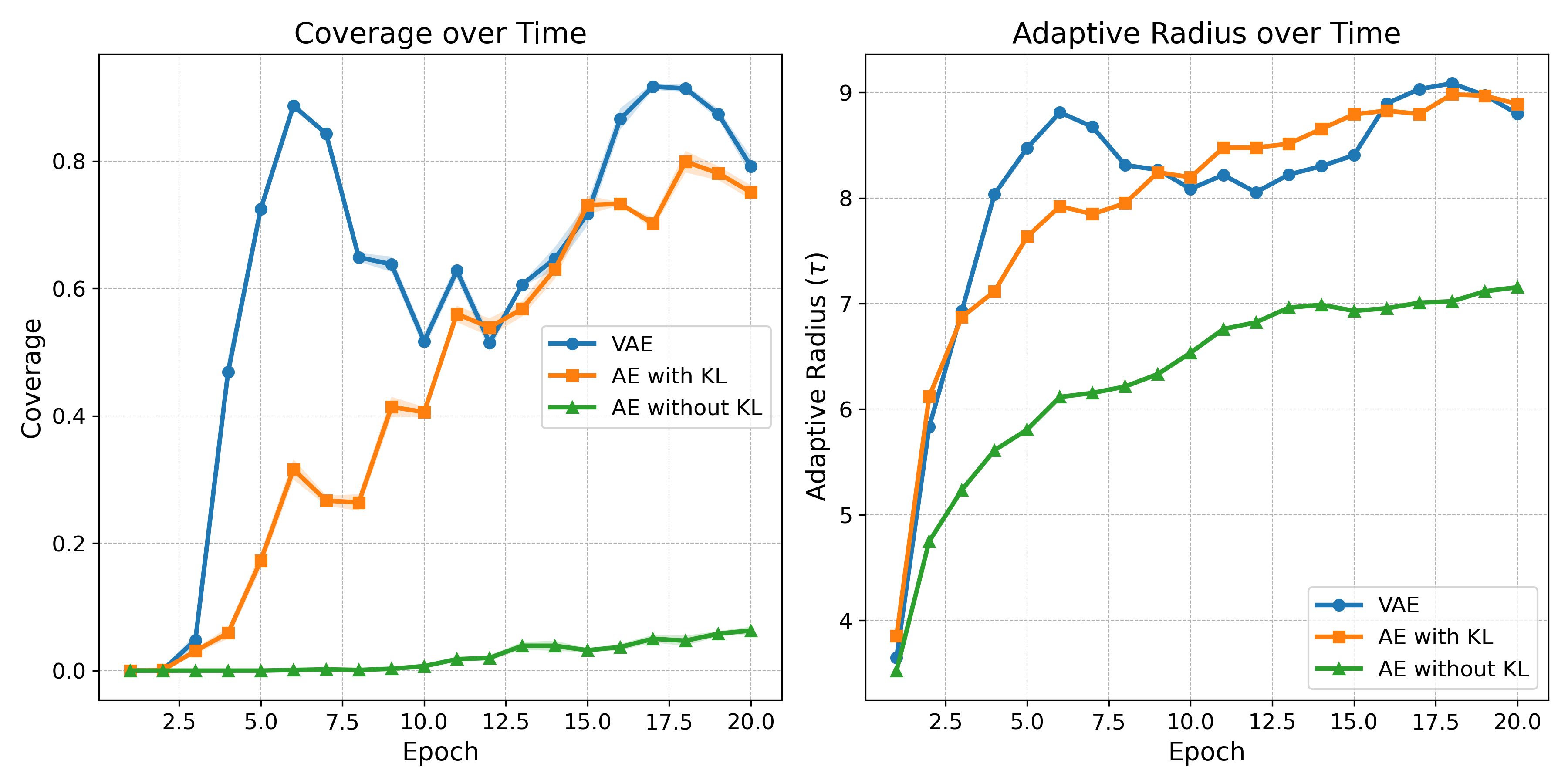} 
    \includegraphics[width=0.32\linewidth]{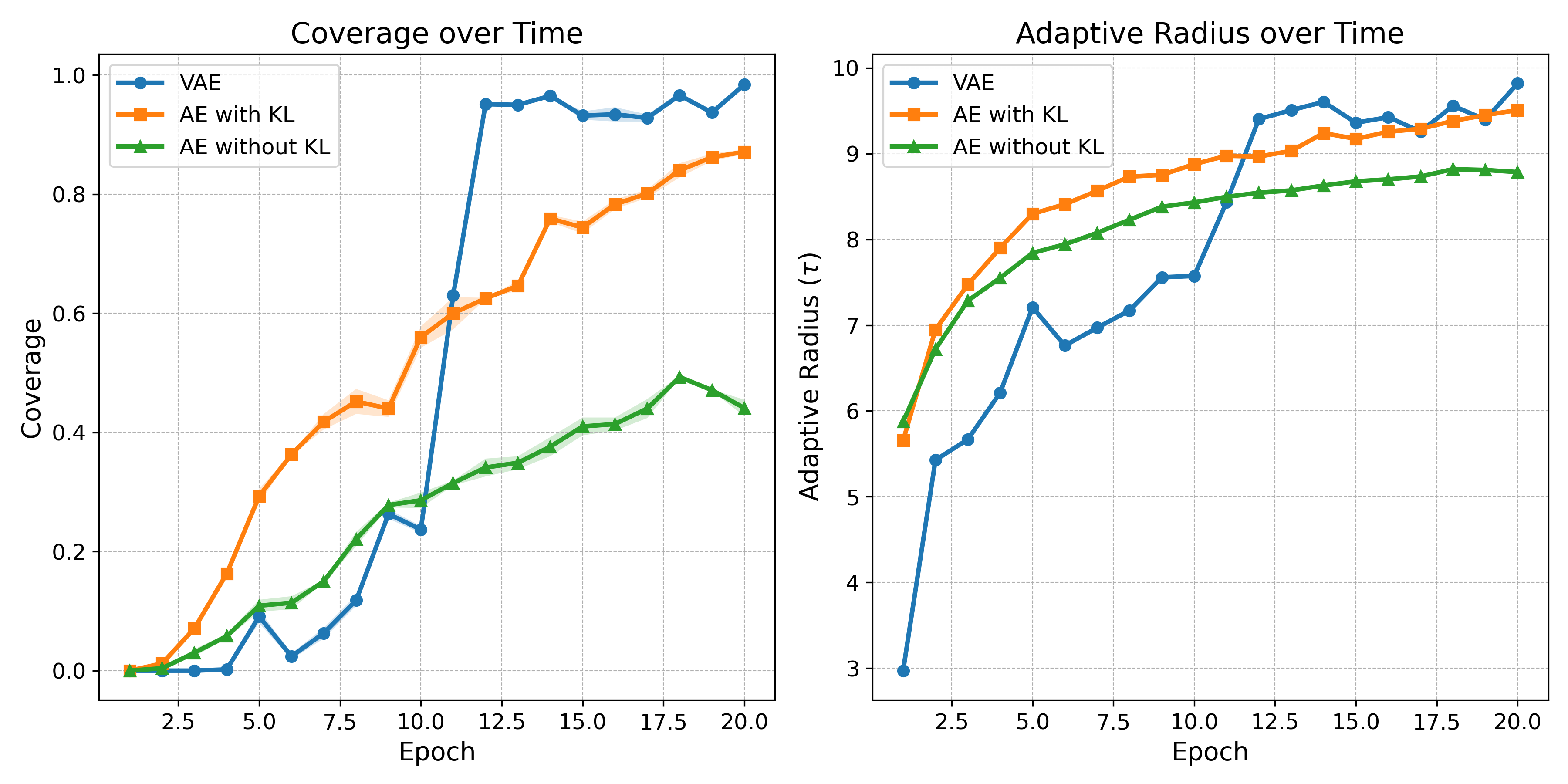}
\end{tabular}
\caption{
Dynamic Latent Coverage and adaptive radius \( \tau \) during training on CelebA (left), FashionMNIST (middle), and MNIST (right). All models use Adam with learning rate 0.001. The CelebA model employs a custom convolutional architecture, while FashionMNIST and MNIST use feedforward networks. Training runs for 10 epochs on CelebA and 20 epochs on the MNIST variants. The hyperparameters follow those defined in \Cref{alg:dynamic_coverage}. Both metrics are computed at the end of each epoch using a fixed evaluation set.
}
\label{fig:coverage_tau_curves}
\end{figure}

As shown in \Cref{fig:coverage_tau_curves}, we conduct experiments to monitor the dynamic changes of the coverage and \( \tau \) metrics during training on three datasets: CelebA, MNIST, and FashionMNIST. Three models are evaluated: standard VAE, AE with KL, and AE without KL (standard AE). We first observe that \( \tau \) changes across epochs, which justifies the necessity of our adaptive radius design. Over the course of training, the coverage increases for all models, but the models with KL regularization exhibit faster growth compared to the standard AE. This indicates that the KL divergence constraint actively contributes to the regularization of the latent space. This effect will be further validated in the experiments presented in the next section.

We now know that KL divergence and reparameterization jointly regularize the latent space, as shown by quantitative analysis using the coefficient of variation (CV) \Cref{fig:cv_plot}. However, it remains unclear whether the correct formation and regularization of semantic manifolds arise specifically from the constraint imposed by KL divergence (abstracted as "compactness of the latent space"), or from the combined effect of KL divergence and reparameterization? This is distinct from the concept of latent space regularization.

In our experiment \Cref{fig:combined_figures}, a separately trained decoder fails to generate via probabilistic sampling, yet still responds to dimensional perturbations, indicating continuity along orthogonal directions in the latent space. Combined with our observation in \Cref{fig:coverage_tau_curves}, that coverage increases whenever KL divergence is applied, progressively filling the semantic manifold.

We can conjecture that the KL divergence itself may shape the geometry of the latent space. 
Understanding this constraint also helps explain why VQVAE succeeds in image generation: despite its discrete latent structure, VQVAE implicitly enforces a similar form of compactness through codebook learning and the commitment loss, effectively regularizing the semantic manifold in a way analogous to KL regularization in continuous VAE.

\subsection{Model Architecture}

\begin{figure}[h!]
    \centering
    \includegraphics[width=0.5\textwidth]{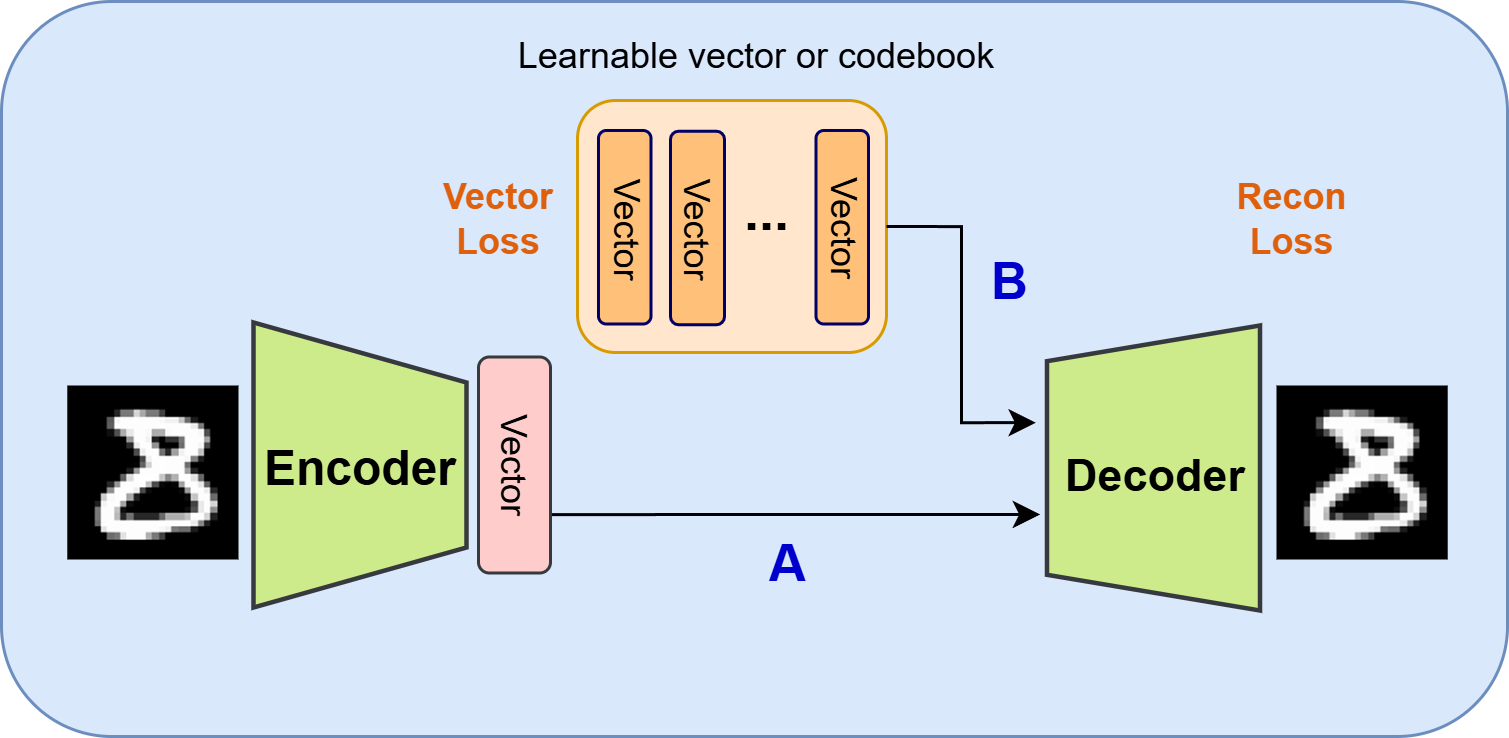}
    \caption{Model Architecture. Path A: for our experiment, where the encoder output is regularized toward the nearest code in a learnable codebook. Path B: VQ-VAE, which uses discrete latent codes. }
    \label{fig:model_architecture}
\end{figure}

As shown in \Cref{fig:model_architecture}, we train an autoencoder with a learnable codebook \( \mathcal{C} = \{c_1, \dots, c_K\} \). For input \( x \), the encoder produces \( \mu = f_{\text{enc}}(x) \), and the nearest code is found as \( k^* = argmin_{k} \| \mu - c_k \|_2 \). The loss combines reconstruction and codebook regularization:
\[
\mathcal{L} = \|x - f_{\text{dec}}(\mu)\|_2^2 + \lambda \|\mu - c_{k^*}\|_2^2.
\]
This encourages the latent space to form compact clusters around learnable centers. While constraining \( \mu \) toward zero would also promote compactness, a learnable codebook allows richer geometric structures. In the figure, path A corresponds to our method, and path B to VQ-VAE, establishing a conceptual connection: both approaches, despite different mechanisms, regularize the latent space to shape semantic manifolds. Notably, VQ-VAE is a special case of our framework, where the latent representation is replaced with the code \( c_{k^*} \) during decoding.

\subsection{Training Autoencoders via Compactness Regularization}

\begin{figure}[h!]
\centering

\begin{subfigure}{0.32\columnwidth}
    \centering
    \includegraphics[width=\linewidth]{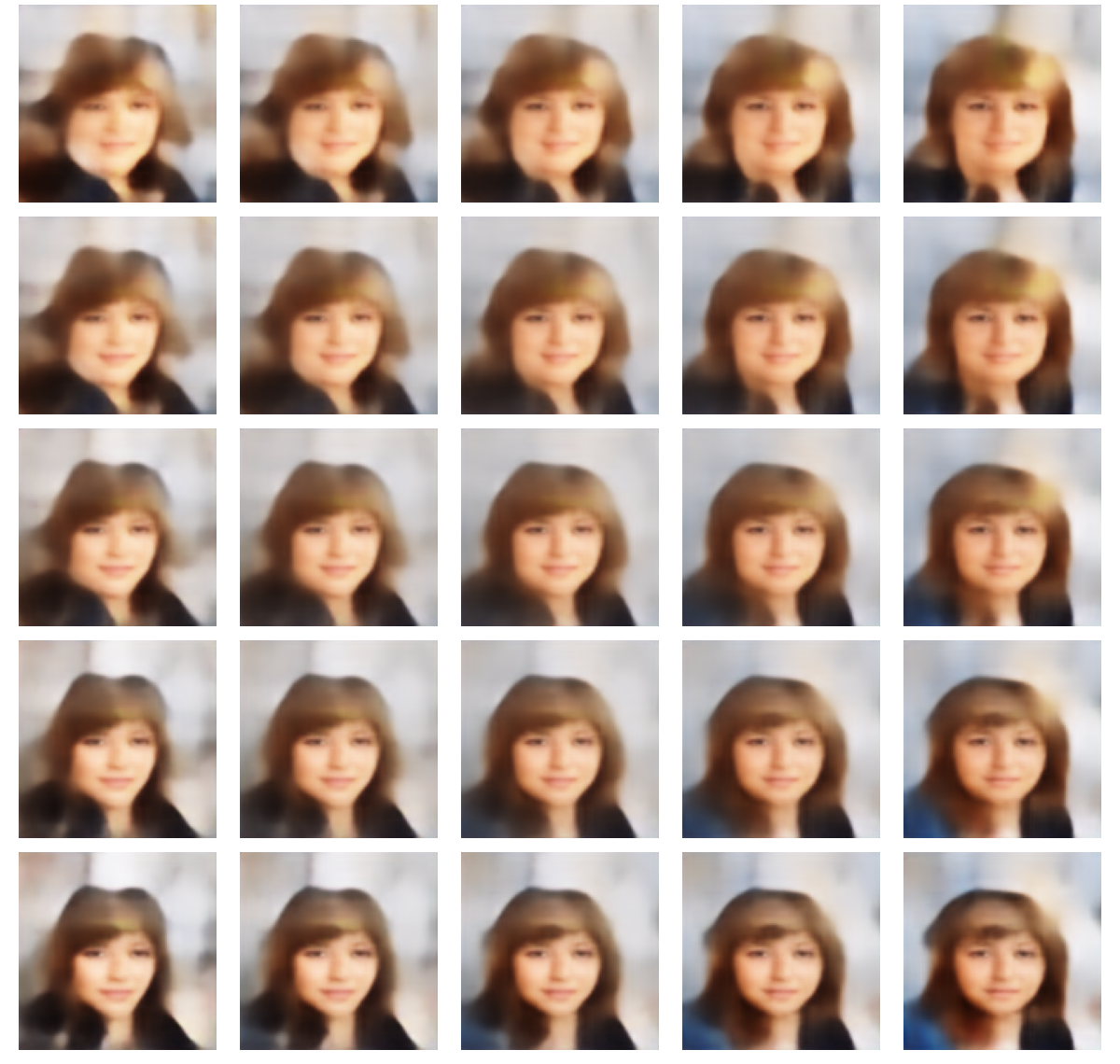}
    \caption{CelebA}
    \label{fig:celeba_perturb}
\end{subfigure}
\begin{subfigure}{0.32\columnwidth}
    \centering
    \includegraphics[width=\linewidth]{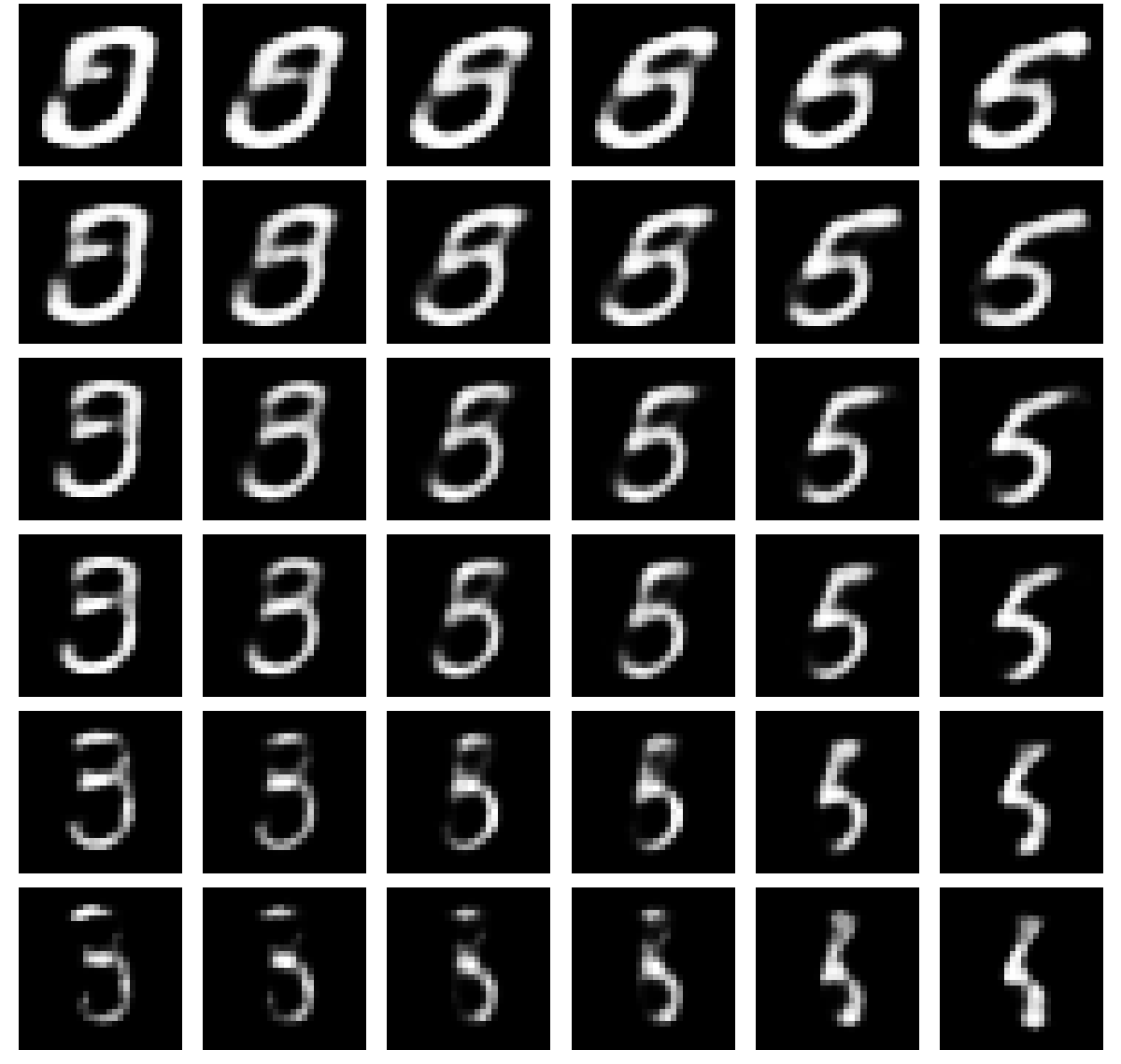}
    \caption{MNIST}
    \label{fig:mnist_perturb}
\end{subfigure}
\begin{subfigure}{0.32\columnwidth}
    \centering
    \includegraphics[width=\linewidth]{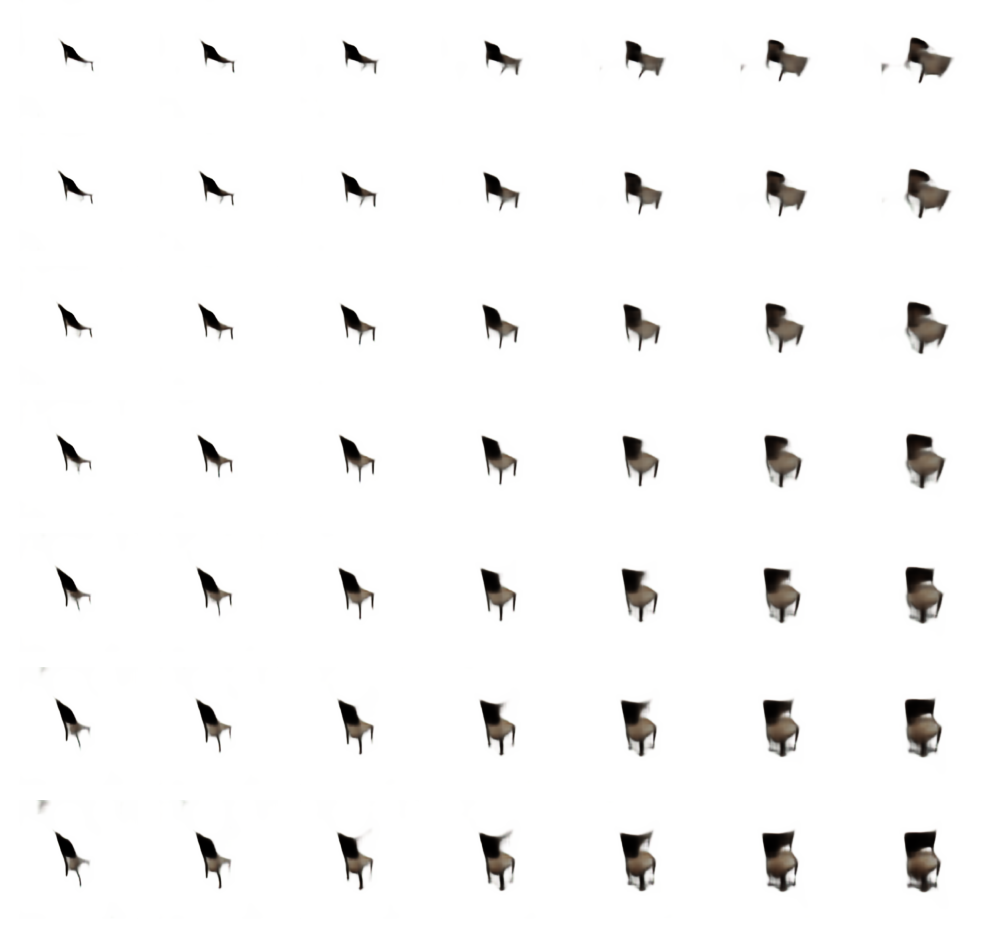}

    \caption{Chairs Dataset~\cite{Aubry14}}
    \label{fig:chairs_perturb}
\end{subfigure}

\begin{subfigure}{0.5\columnwidth}
    \centering
    \includegraphics[width=\linewidth]{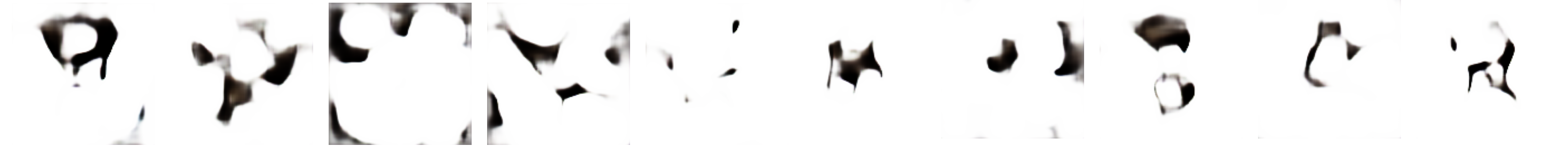} 
    \caption{Chairs Dataset~\cite{Aubry14}, random sample from \( z \sim \mathcal{N}(0, I) \).}
    \label{fig:sampling362}
\end{subfigure}

\caption{
Top row: Dimensional perturbation analysis on autoencoders regularized with a learnable codebook ( \( \lambda  = 0.1 \) ). Each panel shows decoder outputs when perturbing each latent dimension by \([-3, -2, -1, 1, 2, 3]\) (horizontal axis), with dimensions along the vertical axis. Models are trained for 5 epochs using Adam (\( \text{lr} = 0.001 \)). 
Bottom: Generated samples from \( z \sim \mathcal{N}(0, I) \), passed through the decoder. 
}
\label{fig:perturbation_and_sampling}
\end{figure}

As shown in \Cref{fig:perturbation_and_sampling}, the autoencoder with Euclidean codebook regularization exhibits smooth and meaningful responses to dimensional perturbations across all three datasets, consistent with the observations in \Cref{fig:combined_figures}. This suggests that it is the norm regularization induced by the KL divergence, enforcing compactness in the latent space that enables continuous and structured transitions along orthogonal dimensions. When extending the encoder output to multiple vectors and replacing the continuous code with the nearest codebook entry (path B), the model becomes equivalent to VQ-VAE. In this view, we provide an interpretation of why VQ-VAE can generate coherent samples: the key lies in the compactness of the latent space, not necessarily in stochastic quantization. Meanwhile, in the experiment where a separate decoder is trained without access to the original encoder path, sampling fails to produce meaningful outputs, consistent with the findings in  \Cref{fig:combined_figures,fig:revised_vae}. This highlights the role of reparameterization: not as a generative necessity, but as a mechanism to enrich the distribution of latent points during training.

\subsection{Extending the constraint from a single vector to the codebook.}

\begin{figure}[h!]
    \centering
    \includegraphics[width=1\textwidth]{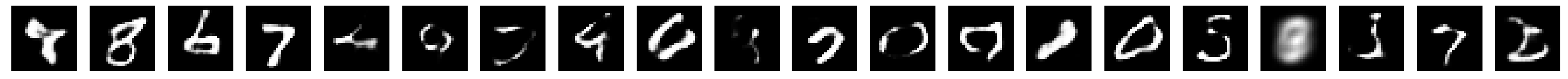}
    \caption {AE constraint experiment on MNIST, trained for 20 epochs with Adam optimizer ( \( lr = 0.001 \) ), extending the constraint from a single vector to the codebook. The figure shows the reconstruction results from 20 codebook vectors directly fed through the decoder.}
    \label{fig:visualize_vector} 
\end{figure}

To further investigate the underlying dynamics and move closer to VQVAE, we extend the constraint from a single vector to the entire codebook. As shown in  \Cref{fig:visualize_vector} , on the MNIST dataset, each codebook entry learns a fragment of an image, serving as a kind of \textbf{"cluster center"}. Each vector captures a piecewise feature. Interestingly, the fourth-to-last vector becomes a blurry average image, clearly failing to learn meaningful content.

\subsection{From Single Code to Codebook: Semantic Space Collapse}

\begin{figure}[h!]
    \centering
    \begin{subfigure}[t]{0.48\linewidth}
        \centering
        \includegraphics[width=\linewidth]{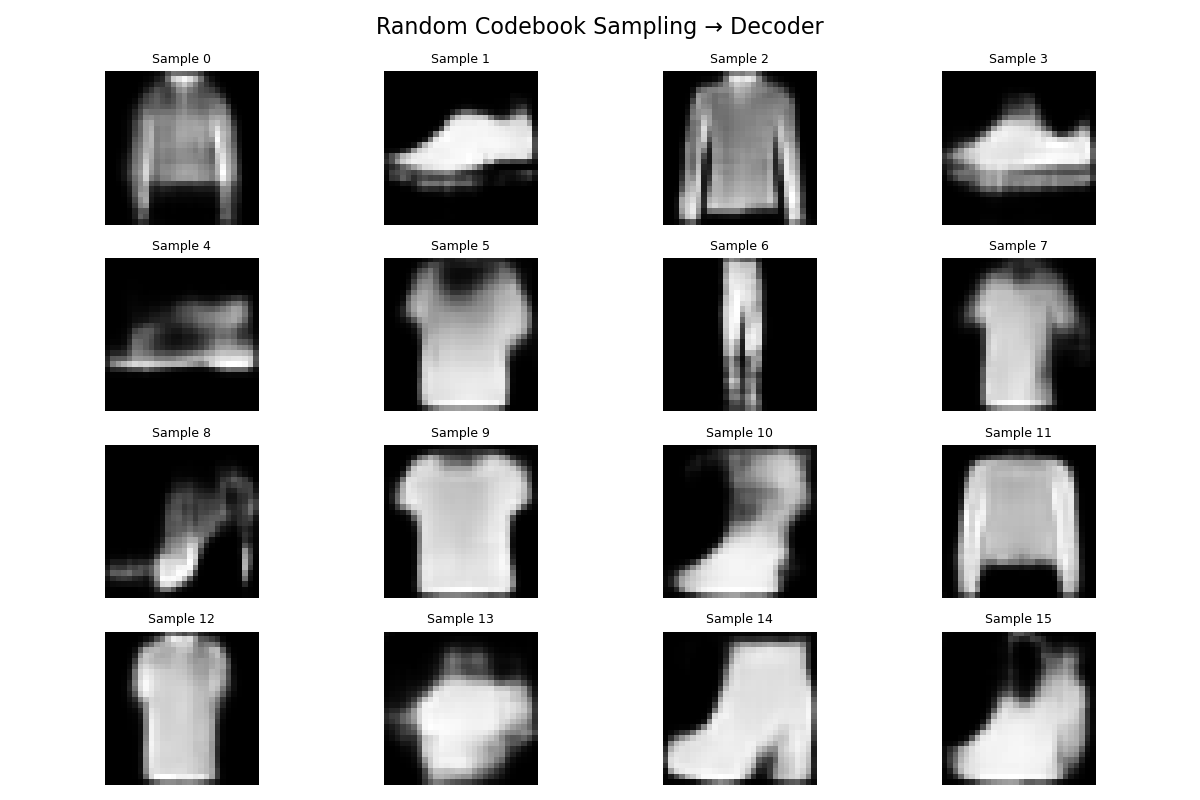}
        \caption{Codebook with 20 vectors, encoder outputs 4 vectors.}
        \label{fig:rs_sub1}
    \end{subfigure}
    \hfill
    \begin{subfigure}[t]{0.48\linewidth}
        \centering
        \includegraphics[width=\linewidth]{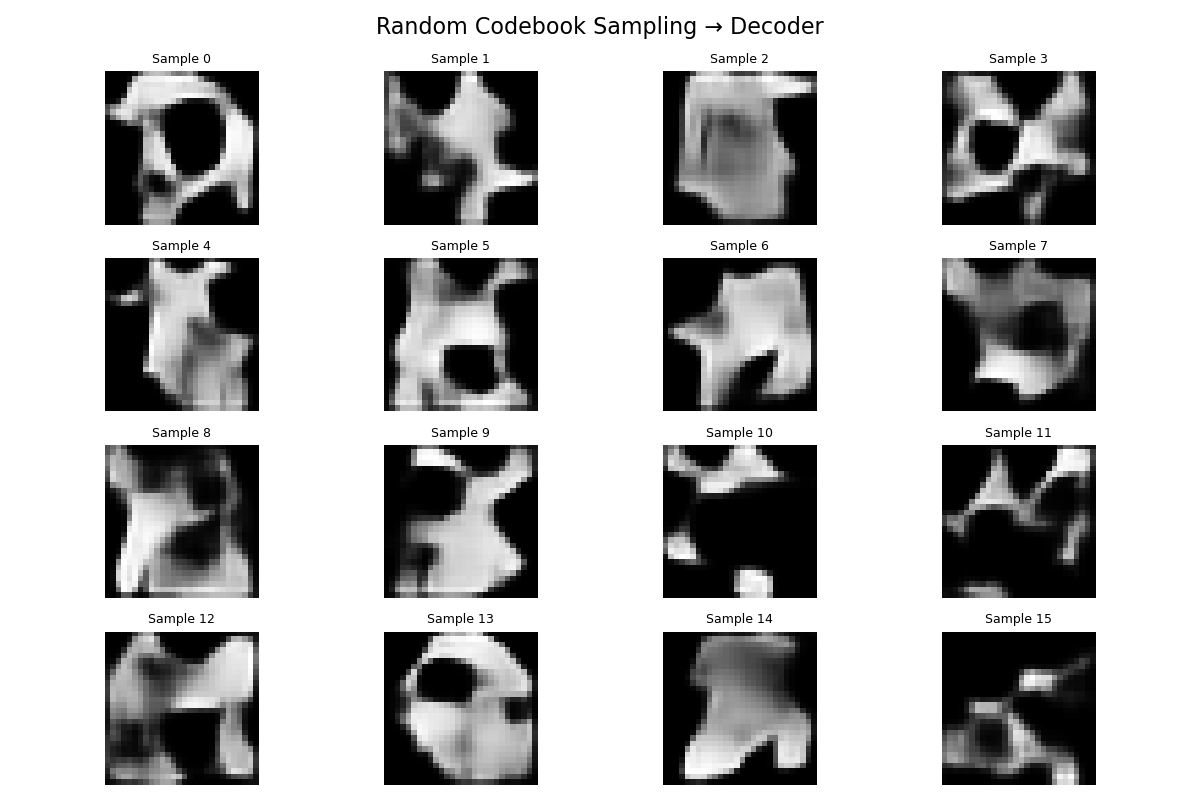}
        \caption{Codebook with 50 vectors, encoder outputs 20 vectors.}
        \label{fig:rs_sub2}
    \end{subfigure}
    \caption{
        Random codebook sampling experiments on FashionMNIST using a convolutional autoencoder trained for 5 epochs with Adam (\(\text{lr} = 0.001\)). 
        The codebook is updated via EMA to avoid collapse to identical vectors. 
        Latent representations follow the VQVAE~\cite{VQVAE} convention: feature maps of shape \([B, C, H, H]\), where each codebook vector corresponds to a channel-wise entry \([B,C, i, j]\) at spatial position \((i,j)\). 
    }
    \label{fig:random_sampling_ablation}
\end{figure}

Next, we demonstrate how the semantic space degenerates. We then increase the number of output vectors from the encoder. As shown in \Cref{fig:random_sampling_ablation}. This "VQVAE-like approximation" degenerates into a deterministic discrete encoder as we scale up the codebook size and the number of encoder output vectors. While it still correctly reconstructs inputs, whether using the encoder’s representation or finding the nearest neighbor in the codebook, randomly sampling from the codebook produces meaningless, fragmented outputs. The model retains only reconstruction capability.  

Although the precise underlying mechanism remains unclear and we cannot yet offer a definitive explanation, we propose a plausible hypothesis: unlike standard VQVAE, our reconstruction path directly passes the encoder’s output to the decoder, while the codebook serves only as a constraint. When the codebook grows large, this creates an "inverse compactness constraint", which means that effectively pulling the semantic space apart and prevents coherent learning.  

Why standard VQVAE does not suffer from this issue may lie in the fact that its decoder receives only codebook vectors, not the raw encoder outputs. This is akin to the "contractual mechanism" between encoder and decoder induced by reparameterization, which we previously discussed, effectively anchoring meaningful points in the latent space. However, this requires further investigation.

\section{Conclusion}

We can conclude that. First, the constraining effect of the KL divergence on the latent space is key to the emergence of semantics, rather than the reparameterization itself. This is because simply constraining the KL divergence and replacing the decoder with a straight-through autoencoder yields the same  performance. However, reparameterization itself serves as a mechanism to form a Gaussian ellipsoid and with the conjunction of KL constraint, regularizes the latent space to be more uniform in terms of CV metrics, thus acting as regularization. Reparameterization, introduces definitional points between the encoder and decoder which enables random sampling. VAE uses KL divergence to force the Gaussian sphere to have a radius of "1". We have also shown that the crucial aspect is preventing the ellipsoid from collapsing, rather than strictly enforcing radius,  allowing it to grow freely can achieve performance identical to VAE. KL divergence serves as a method for latent compactness, we could alternatively use the MSE distance to a learnable codebook as a regularizer. Based on this insight, we propose a unified model that further explains why VQ-VAE possesses generative capability.

While our framework offers a new geometric lens, its formal theoretical underpinnings and a more comprehensive explanation for other phenomena remain an important subject for future investigation.

\section{Related Work}
Early work on linear autoencoders established that they effectively perform PCA when trained with squared reconstruction loss~\cite{AEisPCA}, suggesting that the learned latent space captures the primary modes of the data. This observation was later echoed in VAE, where VAE also recover the principal subspace of the data distribution~\cite{VAEisPCA}, further hinting at a shared geometric mechanism between classical AE and VAE despite their differing probabilistic foundations.

Numerous research explored how standard AE can be endowed with generative capabilities, including regularizing the latent space with topological constraints~\cite{AE_manifold,VALLS,ELASIA}, enforcing smoothness via contractive or denoising objectives~\cite{rifai2011contractive}, or sampling from neighborhoods around encoded points~\cite{bengio2013generalizing}. These efforts reflect a growing understanding that generation in latent space depends not only on stochasticity but also on the structure and regularity of the learned manifold.
Indeed, several works have interpreted VAE through the lens of manifold learning \cite{Use_manifole_in_vaevq01,Use_manifole_in_vaevq02,berg2018dont}. 

More recently, the importance of compactness has been highlighted as a key factor for effective generation. A notable work ~\citet{bortoli2021variational} proposed a new generative model built upon this principle, arguing that compactness, rather than stochasticity enables reliable decoding and interpolation. Our work builds on this insight, but provides a more systematic disentanglement between the roles of compactness and reparameterization, clarifying how geometric regularization through the KL constraint shapes the latent space to support generation.

\clearpage
\nocite{*}

\bibliography{iclr2026_conference}
\bibliographystyle{plainnat}


\clearpage
\appendix

\section{Appendix}

\subsection{Algorithm of Dynamic Latent Coverage}

\begin{algorithm}[h!]
\caption{Dynamic Latent Coverage at Epoch \( t \)}
\label{alg:dynamic_coverage}
\small
\begin{tabular}{@{}l@{}}
\textbf{Input:} Encoder \( f^\mu \), eval set \( \mathcal{D} = \{x_i\}_{i=1}^M \), \( M=1000 \) \\
\phantom{\textbf{Input:}} Prior samples \( N=1000 \), \( k_{\text{nn}}=5 \), percentile \( \eta \) (e.g., 95) \\
\textbf{Output:} \( \text{Coverage}_t \in [0,1] \) \\
\\
1: \textbf{for} \( i = 1 \) to \( M \) \textbf{do} \\
\quad \( \mu_i \gets f^\mu(x_i) \) \\
2: \textbf{end for} \\
3: \( S \gets \{\mu_1, \dots, \mu_M\} \) \\
4: \( D \gets \emptyset \) \\
5: \textbf{for} \( i = 1 \) to \( M \) \textbf{do} \\
\quad Find \( k_{\text{nn}} \) nearest neighbors of \( \mu_i \) in \( S \setminus \{\mu_i\} \) \\
\quad \textbf{for} \( j = 2 \) to \( k_{\text{nn}} \) \textbf{do} \quad // skip closest \\
\qquad \( d \gets \|\mu_i - \mu_{(j)}\|_2 \); Append \( d \) to \( D \) \\
\quad \textbf{end for} \\
6: \textbf{end for} \\
7: \( \tau \gets \mathrm{percentile}(D, \eta) \) \\
8: \( c \gets 0 \) \\
9: \textbf{for} \( i = 1 \) to \( N \) \textbf{do} \\
\quad \( d_{\min} \gets \min_j \|\mu_j - z_i\|_2 \) where \( z_i \sim \mathcal{N}(0,I) \) \\
\quad \textbf{if} \( d_{\min} < \tau \) \textbf{then} \( c \gets c + 1 \) \textbf{end if} \\
10: \textbf{end for} \\
11: \textbf{return} \( c / N \)
\end{tabular}
\end{algorithm}

\subsection{Autoencoders possess generative capabilities}

As the predecessor of Variational Autoencoders (VAE), the Autoencoder (AE) has been regarded in some studies as a nonlinear extension of Principal Component Analysis (PCA).
Similar to classification tasks\cite{classification_manifold}, AE projects data onto a new manifold\cite{AE_manifold},
reshaping the structure of the original distribution. While AE is not conventionally treated as a generative model,
prior work has suggested that it can generate new samples through latent space interpolation.\cite{AE_inter01,AE_inter02}
Similar with the operation in StyleGAN.\cite{Gan_inter03,Gan_inter02}
This section will further explore and validate the generative potential of AEs from this perspective.

\subsection{Perturbation and Interpolation of Latent Codes in Autoencoders}
\begin{figure}[h]
    \centering
    \includegraphics[width=0.4\textwidth]{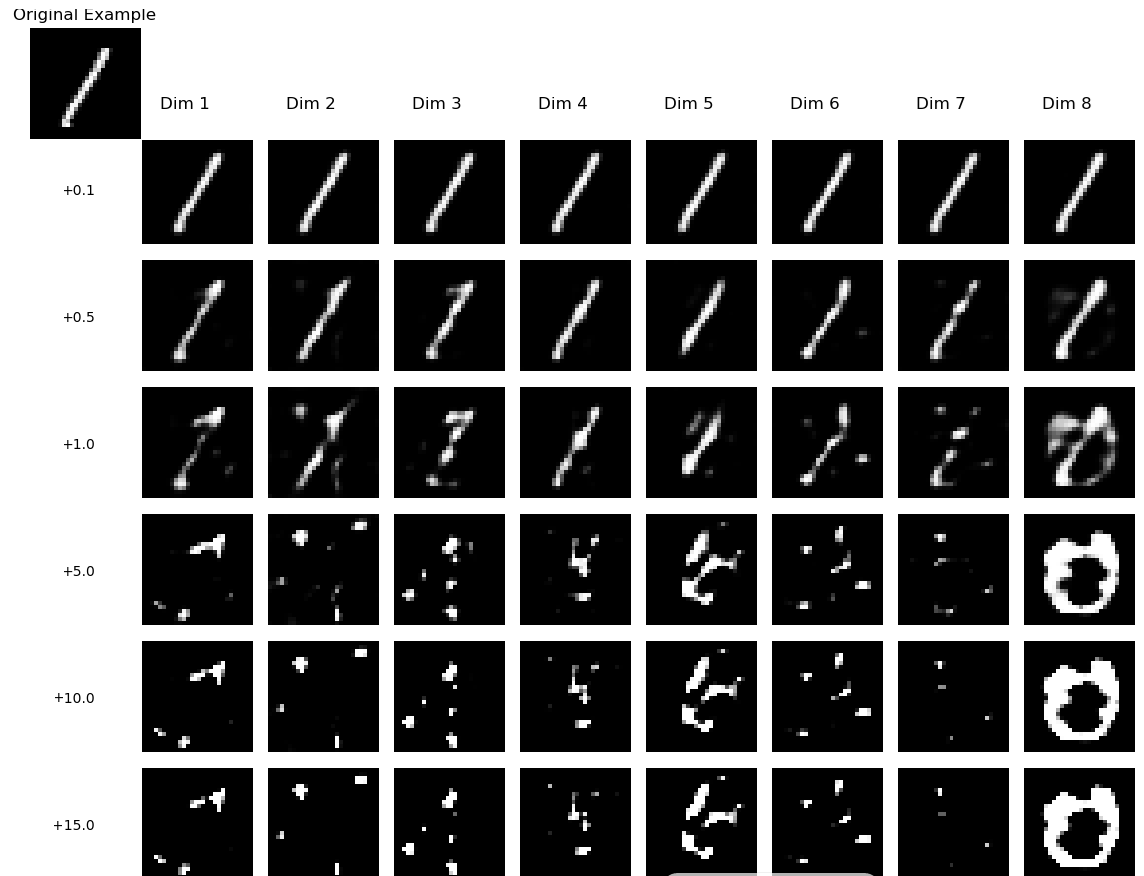} 
    \caption{Add perturbation on the first 8 dimension of the latent space.From +0.1 to +15
Both the encoder and decoder are implemented as 64-dimensional MLPs, 3 layers,with a 128-dimensional latent (encoding) space.} 
    \label{fig1} 
\end{figure}
\begin{figure}[h]
    \centering
    \includegraphics[width=0.8\textwidth]{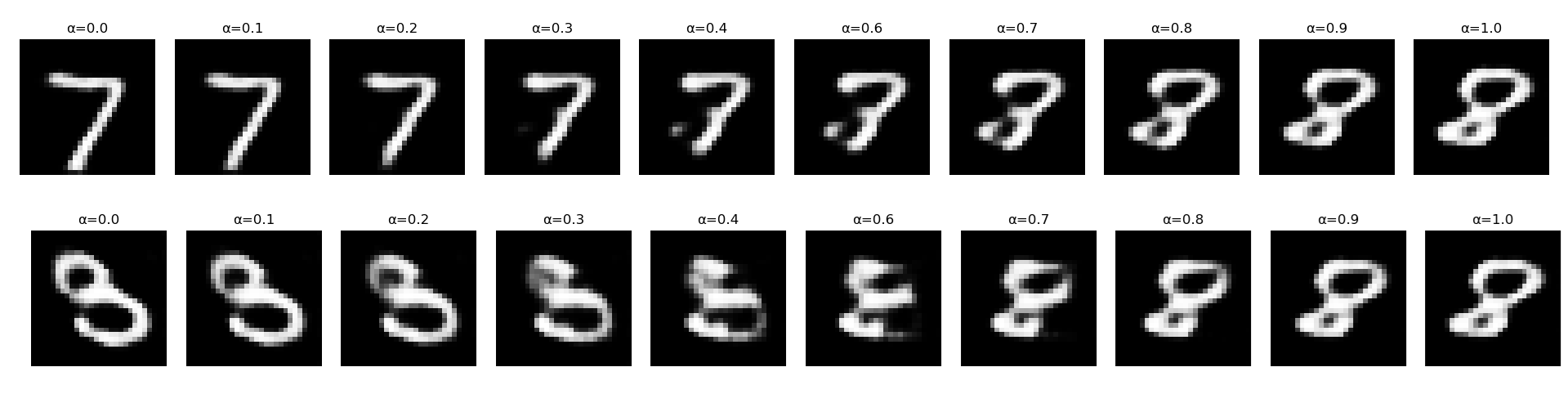} 
    \caption {Linear interpolation of number 7 and 8, 3 and 8.} 
    \label{fig2} 
\end{figure}
Here, we first explore perturbations in the latent space of an autoencoder. As shown in \Cref{fig1}, small perturbations do not significantly alter the decoder's output. However, under large perturbations, the generated images degrade and may even collapse into merely pure black.
Next, we perform linear interpolation between the latent codes of two images \Cref{fig2}. The interpolation is defined as:
\begin{equation*}
\mathbf{z} = (1 - \alpha) \mathbf{z}_1 + \alpha \mathbf{z}_2, \quad \alpha \in [0, 1]
\end{equation*}
where \( \mathbf{z1} \) and \( \mathbf{z2} \)are the latent codes of two input images, and  \( \alpha \) controls the interpolation weight. The decoder produces smoothly transitioning outputs, demonstrating continuous variation between the original images.

\subsection{More Result for Section "Extending the Geometric View: Connection to VQ-VAE" }

\subsubsection{Training With Multiple Vectors.}
\begin{figure}[htbp]
    \centering
    \includegraphics[width=1\textwidth]{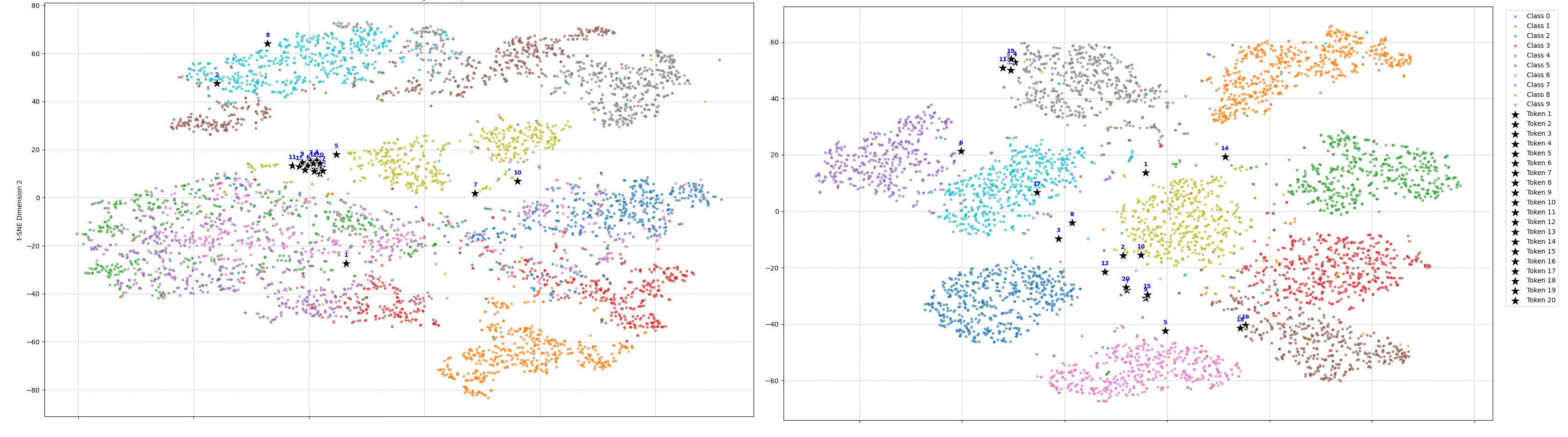}
    \caption {The left image shows a collapse scenario during training on FashionMNIST, while the right image shows normal behavior without collapse on MNIST. These two datasets were chosen to better illustrate the experimental results because FashionMNIST is more prone to collapse, whereas MNIST is difficult to collapse. \textbf{The difference in linear separability for the 10 classes in the two figures is due to the inherent nature of the datasets themselves, and not related to the model.}}
    \label{fig8} 
\end{figure}

On the MNIST/FasionMINIST dataset, we observed that after 20 epochs of training, these vectors landed in different cluster centers. At this point, this constraint was relaxed, and similar to VQ-VAE, degeneration still occurred. Some of these vectors eventually converged to the same location, or simply collapsed into an identical solution, as shown in  \Cref{fig8}

We focus on using a codebook rather than a single vector for training. We previously observed that FashionMNIST is highly prone to collapsp, ultimately becoming indistinguishable from a single vector.

In VQ-VAE, one of the strategies used is Exponential Moving Average (EMA)\cite{VQVAE}. EMA reduces the oscillation of individual vectors through momentum updates, helping to prevent them from collapsing into the same solution. The core formula for updating a codebook vector \( e_k \) using EMA is denoted as follows:
\begin{equation*}
\mathbf{e}_k \leftarrow \mathbf{e}_k + \alpha (\mathbf{z}_q - \mathbf{e}_k)
\end{equation*}

Where:
\( \mathbf{e}_k \) is the\( k \)-th vector in the codebook.
\( \mathbf{z}_q \) is the latent vector.
\( \alpha \) is the momentum (or learning) rate.

\begin{figure}[h!]
    \centering 
    \subfloat[\label{fig11}]{\includegraphics[width=0.4\textwidth]{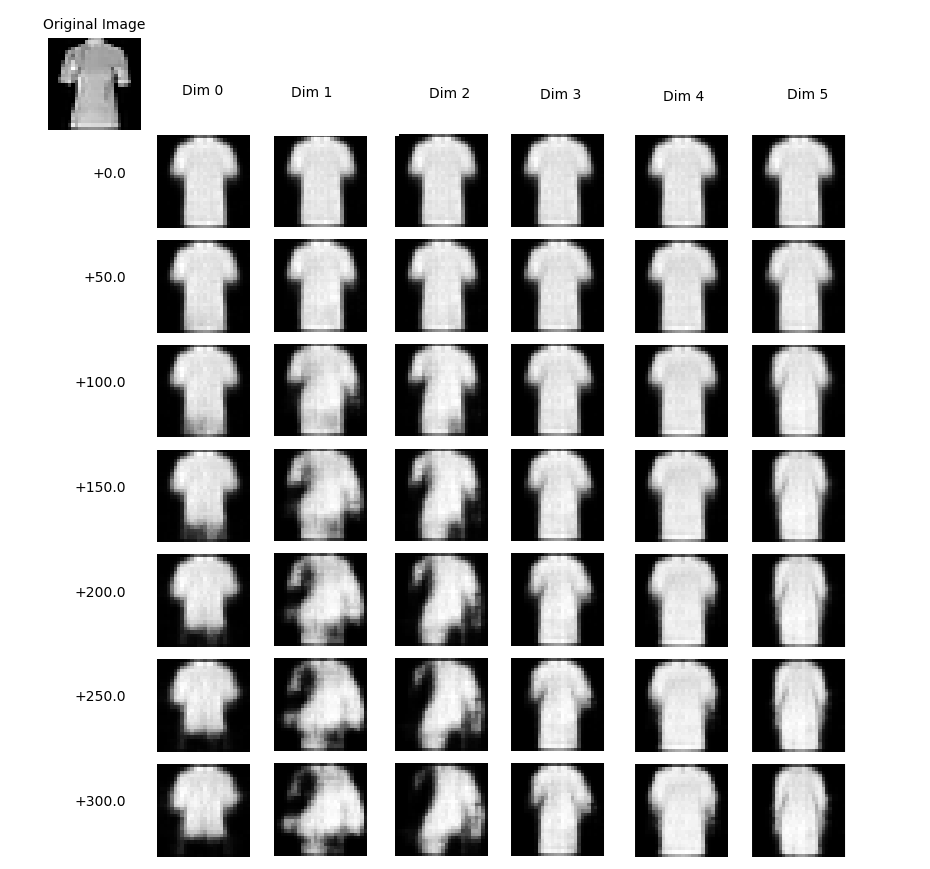}}
    \hfill 
    \subfloat[\label{fig12}]{\includegraphics[width=0.4\textwidth]{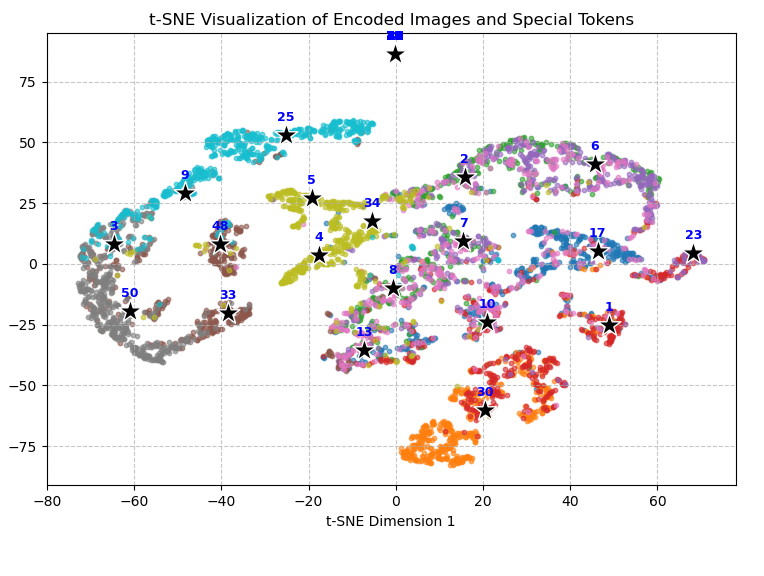}}
    \caption{Experimental results on FashionMNIST.  \protect\subref{fig11} shows the reconstructions from 50 trained vectors, while  \protect\subref{fig12} displays the t-SNE visualization of 20 trained vectors and the encoder output on the test dataset.}
\end{figure}

The final t-SNE is shown in \Cref{fig12}. The data space also becomes more flexible, as depicted in \Cref{fig11}. Here, we applied a +300 additive perturbation, which caused a T-shirt to become longer and its sleeve length to change. Additionally, distortion also occurred.

\subsubsection{Impact of Encoder Capacity on Learnable Vector Quantity}
In  \Cref{fig12}, we observed something interesting: with 50 learnable vectors, only a small number were properly utilized and attracted data.

\begin{figure}[h!]
    \centering
    \includegraphics[width=0.4\textwidth]{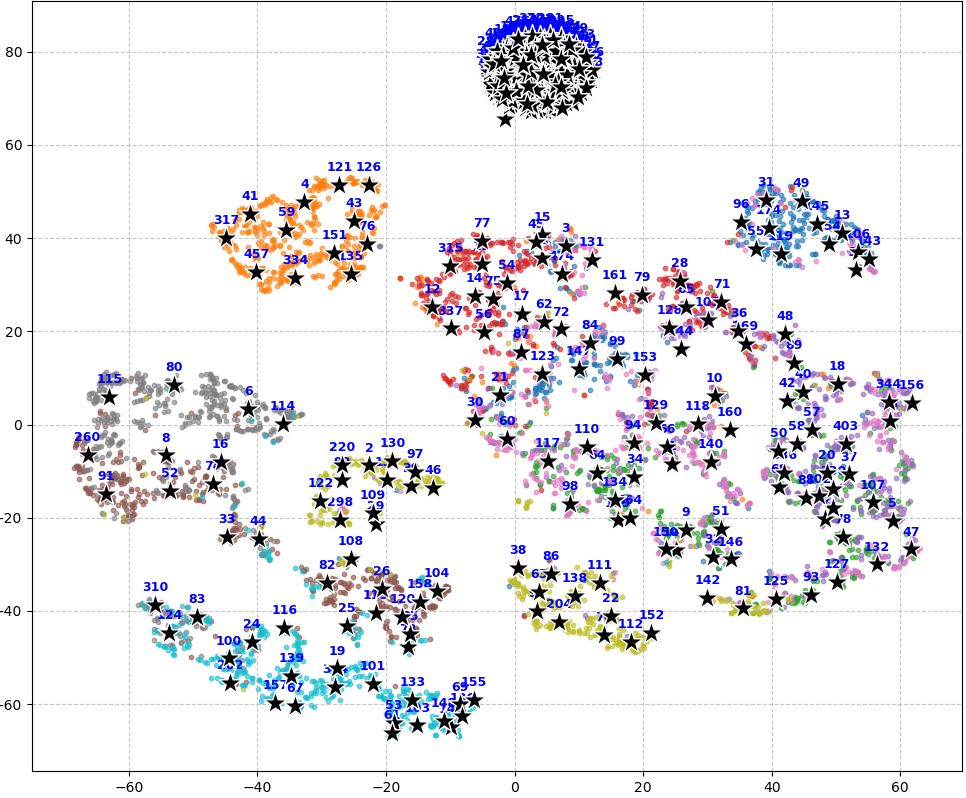}
    \caption {The t-SNE visualization displays the 500 trained vectors and the encoder output from the test dataset. Notably, a large number of these trained vectors still haven't been correctly attracted to data clusters.}
    \label{fig15} 
\end{figure}

\begin{figure}[h!]
    \centering 
    \subfloat[]{\includegraphics[width=0.5\textwidth]{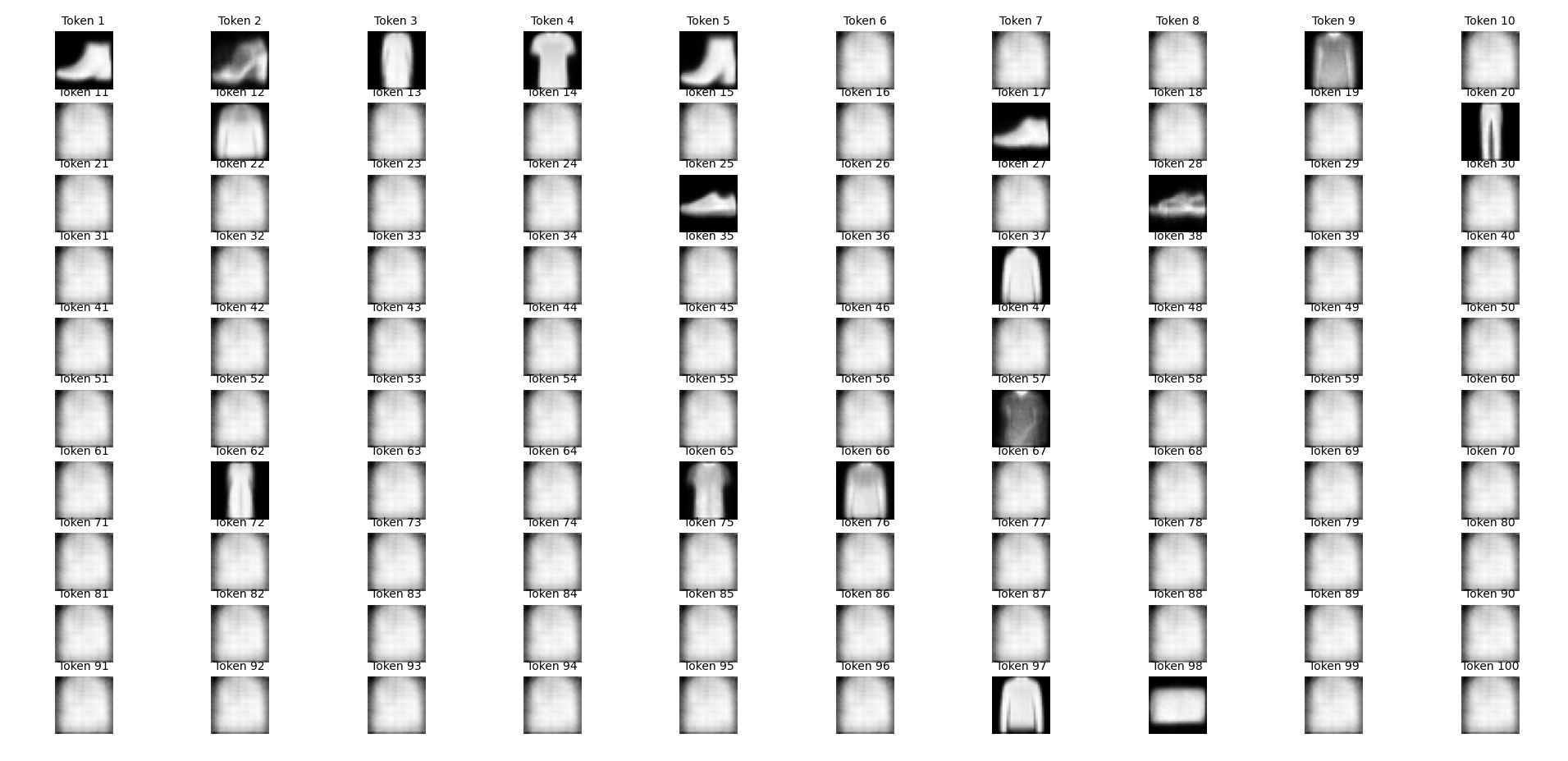}\label{fig:correctly_learned}}
    \hfill 
    \subfloat[]{\includegraphics[width=0.5\textwidth]{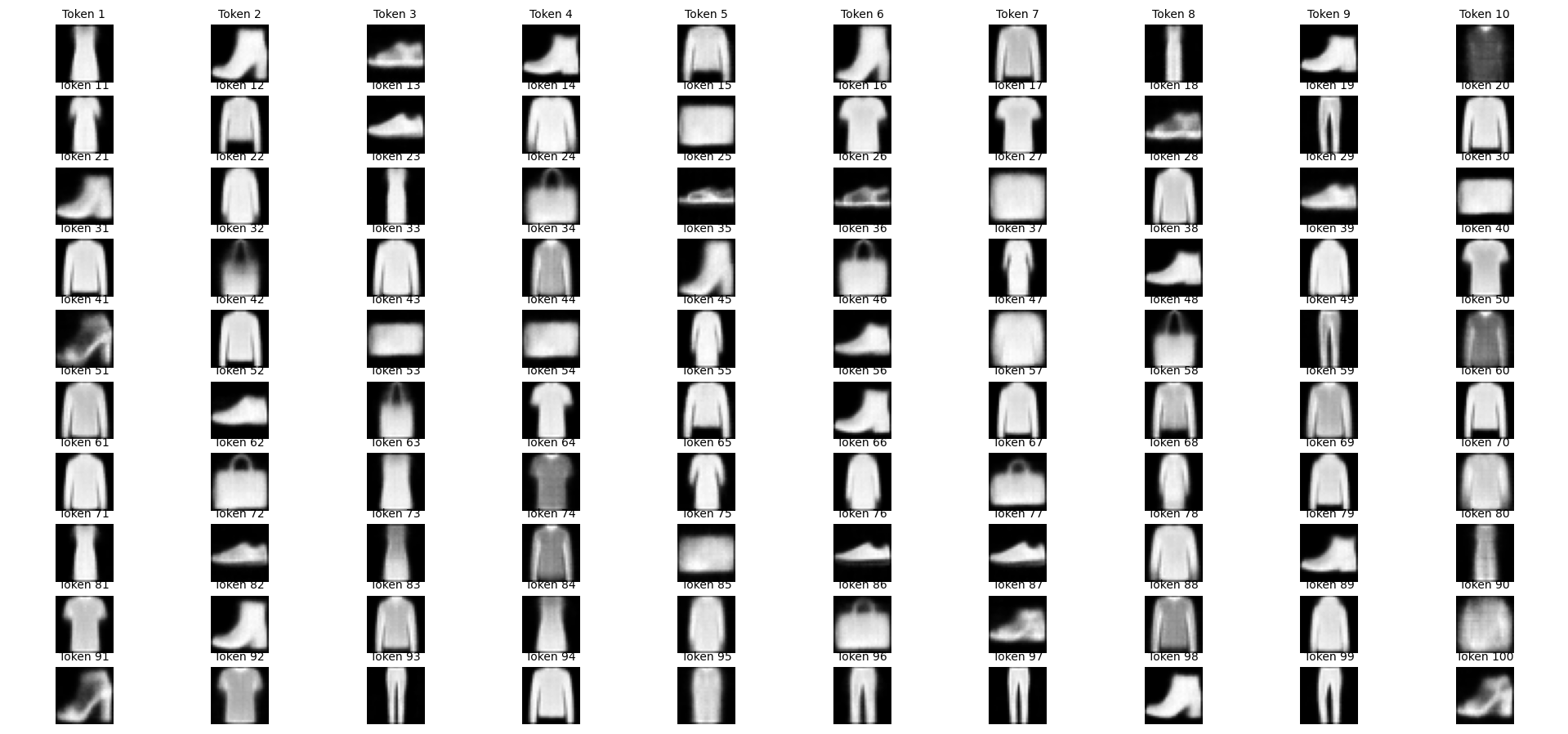}\label{fig:incorrectly_learned}}

    \caption{Experimental results on FashionMNIST with an expanded model.  \protect\subref{fig:correctly_learned} shows reconstructions from randomly selected 100 correctly learned vectors.  \protect\subref{fig:incorrectly_learned} displays reconstructions from many incorrectly learned vectors, where numerous meaningless white ones are present.}
    \label{fig13} 

\end{figure}

To address this, we expanded both the encoder and decoder, making them deeper and incorporating residual connections. The results, shown in \Cref{fig15}, are promising. This time, with 500 vectors, visibly more of them were correctly "attracted" to data clusters. We then randomly selected 200 of these vectors and fed them into the decoder for visualization. \Cref{fig13}

\subsubsection{More Result for Secion "From Single Code to Codebook: Semantic Space Collapse"}

Here we will present additional experiments on "Collapse".

\begin{figure}[h!]
    \centering
    \includegraphics[width=0.55\textwidth]{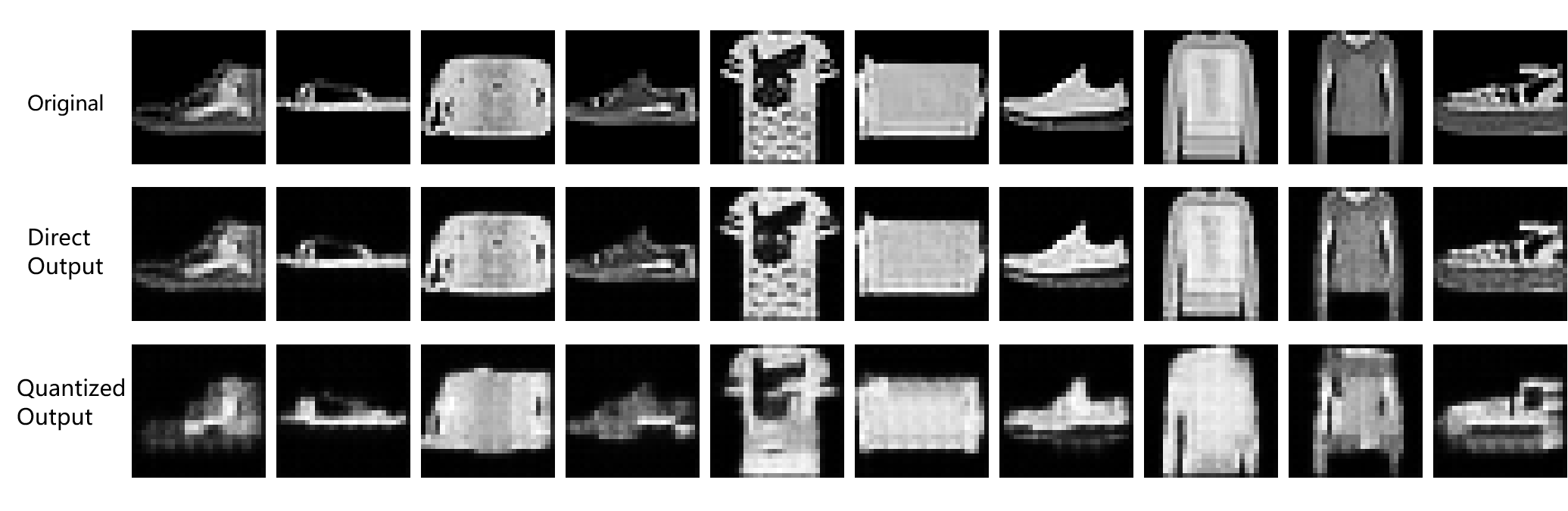}
    \caption {From top to bottom:Original Image,Encoder-Decoder Direct Output,Quantized Output (Codebook to Decoder)}
    \label{fig123} 
\end{figure}
\vspace{-10pt}

\begin{figure}[h!]
    \centering
    \includegraphics[width=0.4\textwidth]{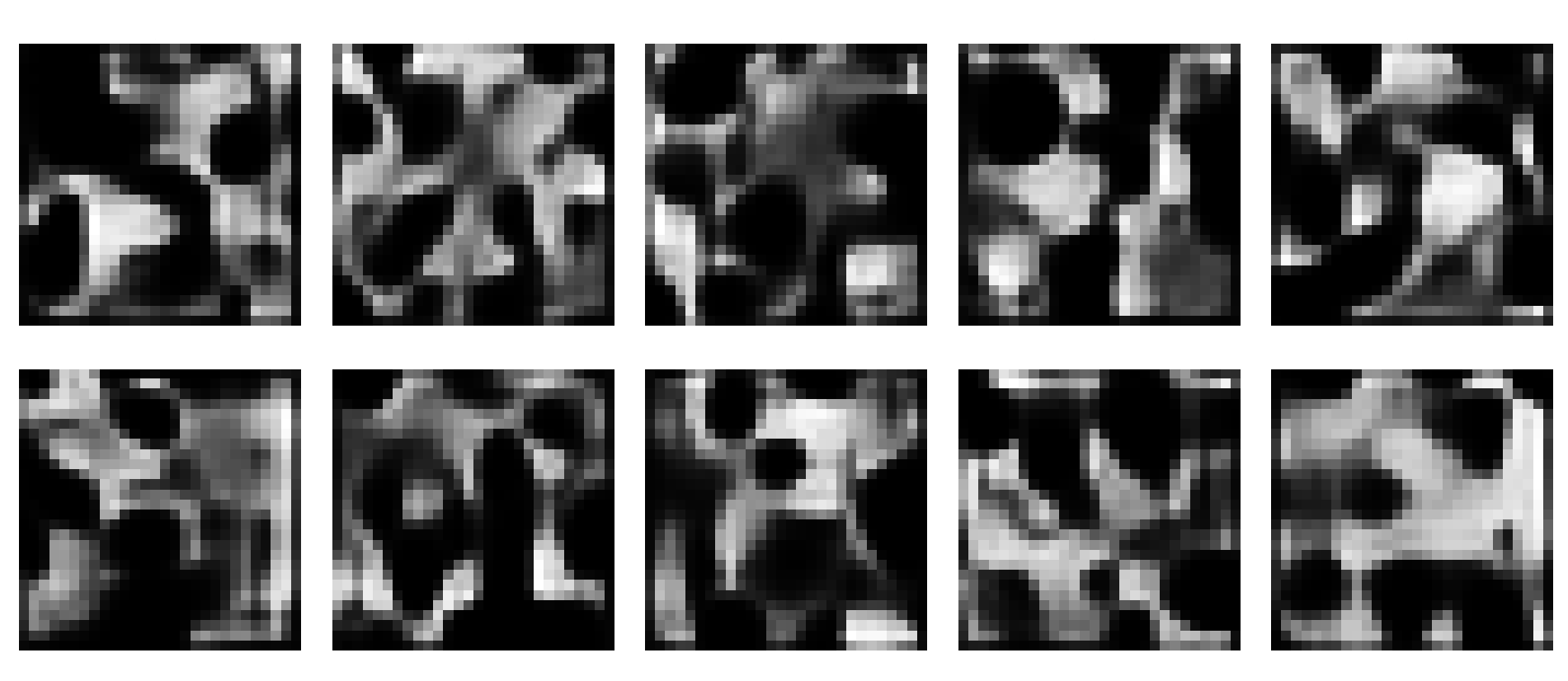}
    \caption {Reconstruction of images generated by randomly combining vectors sampled from the codebook and fed into the decoder.}
    \label{fig456} 
\end{figure}
\vspace{-10pt}

\begin{figure}[h!]
    \centering
    \includegraphics[width=0.7\textwidth]{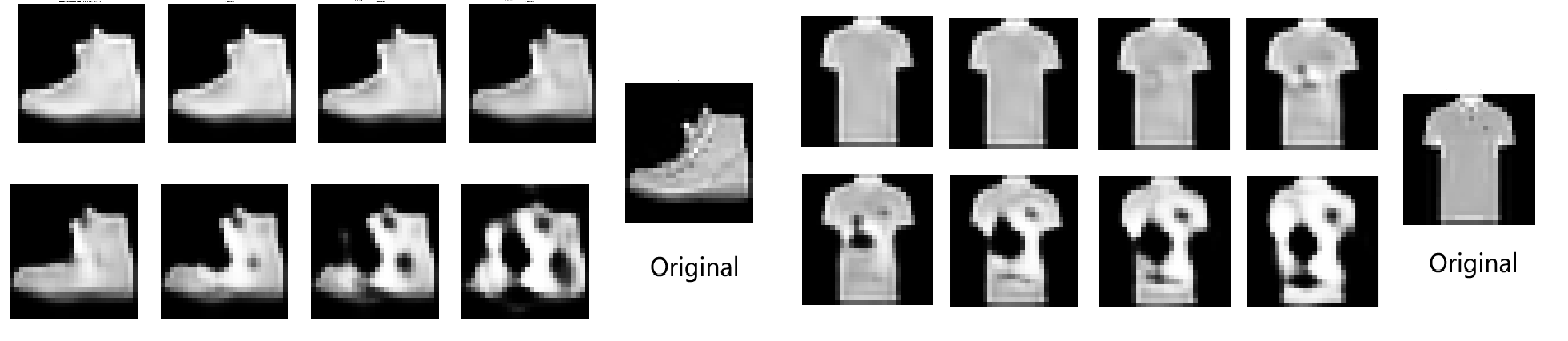}
    \caption {Additive perturbations of +1, +5, +10, +15, +30, +50, and +100 to the first dimension of the first vector in the encoder-to-decoder output.}
    \label{fig789} 
\end{figure}
\vspace{-10pt}
\begin{figure}[h!]
    \centering 
    \subfloat[]{\includegraphics[width=0.5\textwidth]{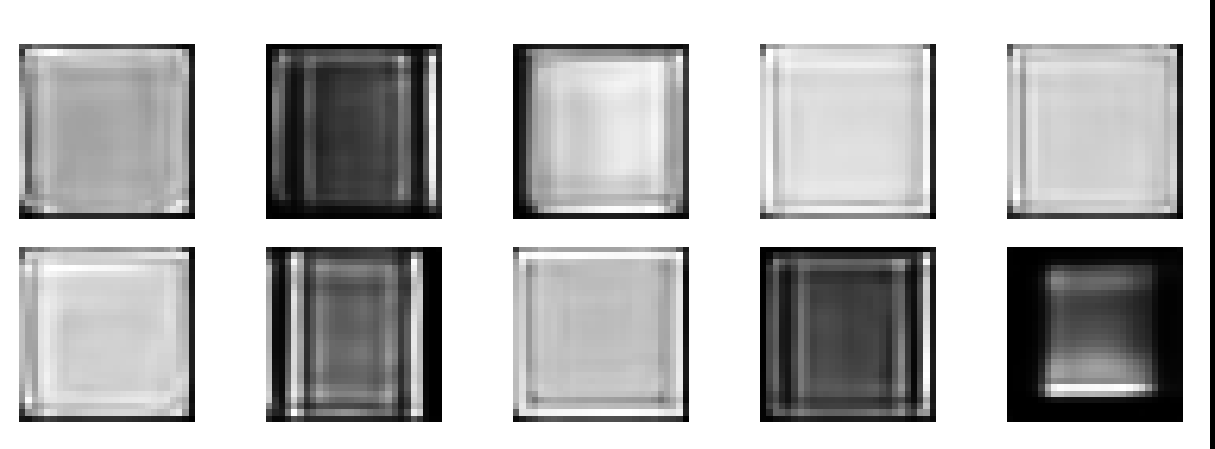}\label{fig:xyz}}
    \hfill 
    \subfloat[]{\includegraphics[width=0.5\textwidth]{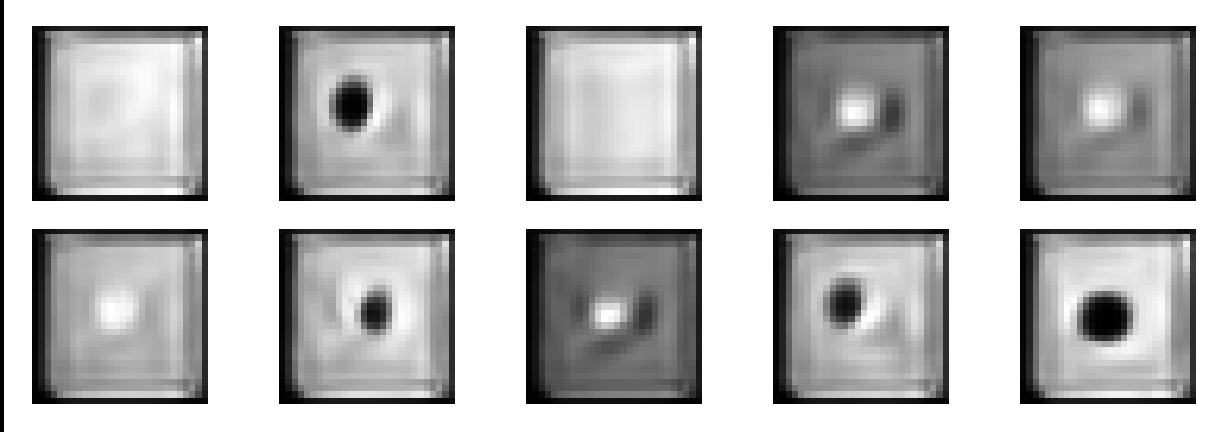}\label{fig:zxy}}

    \caption{Reconstruction effect when codebook vectors are selected and passed to the decoder via broadcasting\protect\subref{fig:xyz} and zero-padding.\protect\subref{fig:zxy}}
    \label{fig159} 

\end{figure}

\vspace{-10pt}

\begin{figure}[h!]
    \centering
    \includegraphics[width=0.6\textwidth]{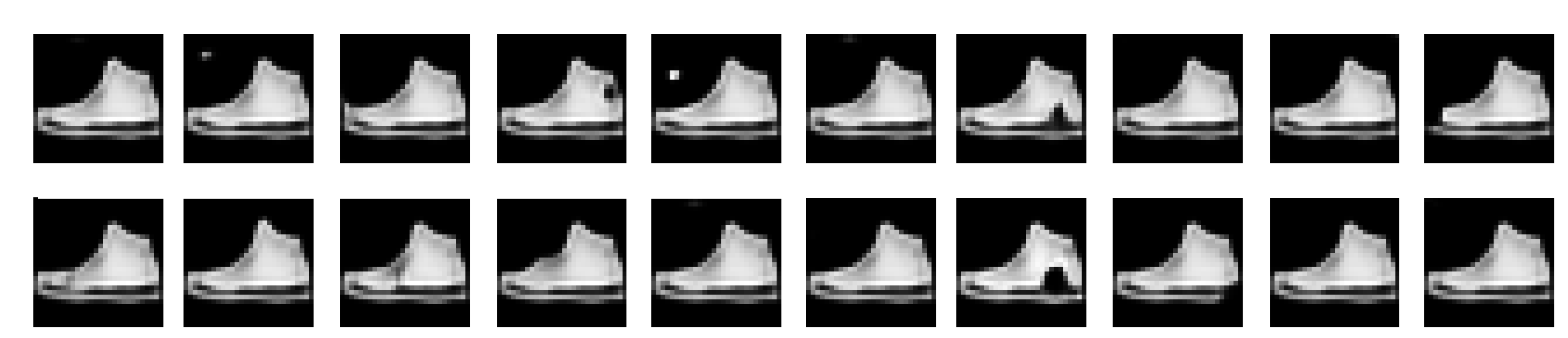}
    \caption {Randomly replace the encoder's encoding with a vector from the codebook.}
    \label{fig753} 
\end{figure}

From \Cref{fig789}, we observe that at this point, no matter how much perturbation is added, the model does not gain a transition capability. Instead, it only increases distortion. In \Cref{fig456}, completely random combinations yield only meaningless results. 
In \Cref{fig753}, arbitrarily replacing one of the vectors with another from the codebook also introduces distortion.
In \Cref{fig159}, the reconstruction results of the codebook vectors themselves are indistinguishable.
In \Cref{fig123}, it is visible that although the model's codebook vectors cannot learn meaningful semantics like in VQ-VAE, they form the original image through correct combinations, and these are indeed random combinations, not a degeneration into an AE. (The codebook indices used for this set of images are in \Cref{code_book_used})

Therefore, we can conclude that in such circumstances, the model tends to become a discrete encoder. Although each vector in the codebook lacks semantic meaning, it can generate relatively clear original images through combination.

\label{code_book_used}
\subsection{Codebook Indices Used in the Experiment}

In our experiment, the encoder's output convolution map has dimensions of 7x7 with 512 channels. Below are the codebook indices matched to each convolutional map location (with channels unfolded as vectors, consistent with VQ-VAE operations).
\[
\begin{bmatrix} 9 & 10 & 10 & 10 & 10 & 10 & 10 \\ 47 & 44 & 44 & 44 & 72 & 64 & 37 \\ 47 & 44 & 44 & 26 & 30 & 96 & 32 \\ 47 & 26 & 72 & 25 & 89 & 11 & 94 \\ 18 & 71 & 69 & 3 & 3 & 1 & 8 \\ 19 & 14 & 20 & 16 & 16 & 16 & 87 \\ 19 & 7 & 7 & 37 & 50 & 37 & 37 \end{bmatrix}
\quad
\begin{bmatrix} 9 & 17 & 21 & 31 & 36 & 79 & 10 \\ 47 & 58 & 24 & 60 & 3 & 95 & 58 \\ 47 & 44 & 54 & 89 & 23 & 37 & 58 \\ 47 & 44 & 59 & 60 & 57 & 38 & 58 \\ 47 & 44 & 59 & 60 & 74 & 38 & 58 \\ 47 & 44 & 59 & 60 & 57 & 38 & 58 \\ 19 & 7 & 54 & 94 & 57 & 38 & 7 \end{bmatrix}
\quad
\begin{bmatrix} 9 & 17 & 21 & 80 & 31 & 79 & 17 \\ 19 & 25 & 0 & 65 & 65 & 0 & 6 \\ 91 & 40 & 65 & 0 & 0 & 89 & 53 \\ 91 & 40 & 65 & 0 & 0 & 89 & 93 \\ 28 & 69 & 69 & 0 & 0 & 33 & 81 \\ 28 & 1 & 69 & 65 & 0 & 42 & 2 \\ 91 & 32 & 33 & 1 & 0 & 33 & 81 \end{bmatrix}
\]

\[
\begin{bmatrix} 9 & 17 & 85 & 36 & 36 & 79 & 10 \\ 47 & 26 & 27 & 0 & 22 & 64 & 58 \\ 47 & 26 & 27 & 0 & 0 & 64 & 58 \\ 47 & 26 & 27 & 41 & 11 & 64 & 58 \\ 47 & 26 & 30 & 66 & 11 & 6 & 58 \\ 47 & 26 & 30 & 81 & 0 & 6 & 58 \\ 19 & 26 & 30 & 93 & 80 & 6 & 7 \end{bmatrix}
\quad
\begin{bmatrix} 9 & 17 & 17 & 10 & 17 & 79 & 10 \\ 19 & 29 & 23 & 82 & 59 & 32 & 6 \\ 91 & 43 & 3 & 3 & 65 & 3 & 66 \\ 73 & 60 & 68 & 65 & 65 & 65 & 74 \\ 73 & 60 & 68 & 68 & 65 & 65 & 94 \\ 73 & 60 & 68 & 60 & 68 & 65 & 94 \\ 76 & 39 & 39 & 39 & 39 & 39 & 75 \end{bmatrix}
\quad
\begin{bmatrix} 9 & 45 & 77 & 36 & 36 & 77 & 17 \\ 19 & 35 & 35 & 35 & 95 & 35 & 50 \\ 47 & 70 & 62 & 35 & 35 & 35 & 50 \\ 47 & 26 & 35 & 87 & 87 & 64 & 58 \\ 47 & 26 & 35 & 87 & 87 & 64 & 58 \\ 47 & 7 & 35 & 87 & 87 & 35 & 58 \\ 19 & 72 & 35 & 87 & 87 & 87 & 37 \end{bmatrix}
\]

\[
\begin{bmatrix} 9 & 10 & 45 & 36 & 5 & 10 & 10 \\ 47 & 44 & 72 & 80 & 66 & 58 & 58 \\ 47 & 44 & 72 & 0 & 66 & 58 & 58 \\ 47 & 44 & 70 & 0 & 49 & 37 & 58 \\ 47 & 44 & 29 & 65 & 32 & 38 & 58 \\ 47 & 26 & 30 & 65 & 3 & 6 & 58 \\ 19 & 7 & 40 & 1 & 1 & 34 & 7 \end{bmatrix}
\quad
\begin{bmatrix} 9 & 10 & 17 & 77 & 77 & 17 & 10 \\ 47 & 44 & 72 & 66 & 4 & 6 & 58 \\ 47 & 26 & 29 & 11 & 84 & 49 & 37 \\ 19 & 59 & 3 & 65 & 65 & 3 & 66 \\ 91 & 67 & 3 & 65 & 89 & 3 & 81 \\ 28 & 55 & 3 & 65 & 65 & 65 & 2 \\ 19 & 25 & 94 & 94 & 96 & 96 & 93 \end{bmatrix}
\quad
\begin{bmatrix} 9 & 10 & 10 & 10 & 10 & 10 & 10 \\ 47 & 44 & 44 & 26 & 7 & 44 & 26 \\ 47 & 26 & 7 & 71 & 67 & 86 & 71 \\ 18 & 71 & 84 & 0 & 0 & 0 & 63 \\ 51 & 80 & 0 & 56 & 63 & 63 & 22 \\ 19 & 50 & 12 & 12 & 14 & 14 & 14 \\ 19 & 7 & 7 & 7 & 7 & 7 & 7 \end{bmatrix}
\]

\[
\begin{bmatrix} 9 & 10 & 21 & 80 & 49 & 17 & 10 \\ 47 & 26 & 4 & 22 & 80 & 6 & 58 \\ 47 & 44 & 4 & 32 & 80 & 38 & 58 \\ 47 & 44 & 70 & 32 & 80 & 38 & 58 \\ 47 & 44 & 72 & 32 & 22 & 38 & 58 \\ 47 & 44 & 72 & 32 & 27 & 38 & 58 \\ 19 & 7 & 72 & 49 & 32 & 38 & 7 \end{bmatrix}
\quad
\begin{bmatrix} 9 & 10 & 10 & 10 & 10 & 17 & 45 \\ 47 & 44 & 26 & 86 & 71 & 84 & 75 \\ 47 & 26 & 4 & 1 & 60 & 55 & 78 \\ 19 & 4 & 63 & 75 & 3 & 22 & 80 \\ 67 & 56 & 8 & 56 & 22 & 32 & 80 \\ 73 & 80 & 80 & 80 & 62 & 27 & 22 \\ 91 & 12 & 14 & 14 & 50 & 14 & 46 \end{bmatrix}
\quad
\begin{bmatrix} 9 & 45 & 36 & 22 & 27 & 31 & 17 \\ 19 & 29 & 11 & 0 & 0 & 1 & 34 \\ 19 & 27 & 11 & 63 & 22 & 0 & 81 \\ 91 & 69 & 11 & 22 & 22 & 0 & 2 \\ 28 & 0 & 65 & 69 & 80 & 0 & 49 \\ 28 & 0 & 65 & 0 & 0 & 0 & 32 \\ 76 & 74 & 39 & 39 & 24 & 63 & 32 \end{bmatrix}
\]

\[
\begin{bmatrix} 9 & 17 & 85 & 13 & 83 & 79 & 10 \\ 47 & 59 & 3 & 68 & 68 & 23 & 38 \\ 47 & 29 & 74 & 65 & 65 & 74 & 6 \\ 19 & 30 & 68 & 65 & 65 & 60 & 34 \\ 19 & 43 & 60 & 68 & 68 & 89 & 53 \\ 91 & 40 & 69 & 60 & 60 & 42 & 93 \\ 91 & 43 & 46 & 46 & 46 & 30 & 53 \end{bmatrix}
\quad
\begin{bmatrix} 9 & 17 & 17 & 17 & 17 & 17 & 17 \\ 19 & 4 & 87 & 67 & 67 & 78 & 78 \\ 91 & 27 & 22 & 22 & 0 & 0 & 0 \\ 76 & 22 & 22 & 0 & 0 & 0 & 0 \\ 73 & 27 & 80 & 0 & 0 & 0 & 80 \\ 73 & 22 & 22 & 63 & 22 & 22 & 24 \\ 91 & 37 & 50 & 50 & 37 & 37 & 37 \end{bmatrix}
\quad
\begin{bmatrix} 9 & 10 & 10 & 10 & 10 & 10 & 10 \\ 47 & 44 & 44 & 7 & 7 & 26 & 7 \\ 47 & 26 & 7 & 25 & 80 & 78 & 66 \\ 19 & 70 & 80 & 30 & 0 & 0 & 16 \\ 18 & 43 & 63 & 32 & 63 & 1 & 80 \\ 51 & 46 & 20 & 20 & 46 & 16 & 46 \\ 19 & 7 & 7 & 7 & 7 & 7 & 7 \end{bmatrix}
\]

\[
\begin{bmatrix} 9 & 45 & 85 & 13 & 83 & 5 & 17 \\ 19 & 43 & 3 & 65 & 8 & 3 & 34 \\ 91 & 56 & 3 & 68 & 65 & 8 & 53 \\ 47 & 50 & 69 & 68 & 68 & 93 & 50 \\ 47 & 58 & 69 & 68 & 68 & 93 & 58 \\ 47 & 72 & 69 & 68 & 68 & 93 & 58 \\ 19 & 72 & 1 & 94 & 94 & 93 & 7 \end{bmatrix}
\quad
\begin{bmatrix} 9 & 10 & 10 & 10 & 10 & 10 & 10 \\ 47 & 44 & 44 & 26 & 86 & 37 & 86 \\ 47 & 44 & 26 & 72 & 48 & 69 & 32 \\ 19 & 26 & 86 & 87 & 22 & 89 & 0 \\ 18 & 87 & 48 & 80 & 80 & 1 & 80 \\ 51 & 14 & 20 & 20 & 20 & 16 & 20 \\ 19 & 7 & 7 & 7 & 7 & 7 & 7 \end{bmatrix}
\quad
\begin{bmatrix} 9 & 17 & 17 & 17 & 17 & 17 & 17 \\ 18 & 71 & 71 & 71 & 71 & 71 & 78 \\ 80 & 0 & 0 & 0 & 0 & 0 & 1 \\ 40 & 69 & 8 & 69 & 69 & 8 & 96 \\ 69 & 0 & 0 & 0 & 0 & 0 & 1 \\ 69 & 1 & 0 & 1 & 1 & 1 & 1 \\ 51 & 14 & 14 & 14 & 12 & 12 & 12 \end{bmatrix}
\]

\[
\begin{bmatrix} 9 & 10 & 17 & 56 & 2 & 17 & 10 \\ 47 & 44 & 72 & 89 & 57 & 37 & 58 \\ 47 & 44 & 72 & 89 & 57 & 38 & 58 \\ 47 & 44 & 70 & 65 & 1 & 6 & 58 \\ 47 & 44 & 59 & 65 & 0 & 6 & 58 \\ 47 & 44 & 54 & 65 & 74 & 38 & 58 \\ 19 & 7 & 54 & 1 & 57 & 38 & 7 \end{bmatrix}
\quad
\begin{bmatrix} 9 & 10 & 45 & 13 & 36 & 5 & 10 \\ 47 & 44 & 72 & 89 & 60 & 53 & 58 \\ 47 & 44 & 54 & 89 & 68 & 34 & 58 \\ 47 & 44 & 54 & 60 & 60 & 34 & 58 \\ 47 & 44 & 59 & 60 & 68 & 53 & 58 \\ 47 & 44 & 29 & 60 & 60 & 53 & 58 \\ 19 & 7 & 25 & 16 & 16 & 12 & 58 \end{bmatrix}
\]


\end{document}